\documentclass[11pt,a4paper]{article}

\usepackage{booktabs}
\usepackage{amsmath}
\usepackage{amssymb}
\usepackage{subcaption}
\usepackage{acronym}
\usepackage{longtable}
\usepackage{tikz}
\usepackage{hyperref}
\usepackage{cleveref}
\usepackage[customfonts]{nxai}
\usepackage{ml_defs}

\usepackage{float}
\usepackage{array}
\usepackage{orcidlink}

\iflatexml
  \renewcommand{\Cref}[1]{\nameCref{#1}~\ref{#1}}
\fi

\newcommand{\timeseries}{time series}

\newcommand{\affmark}[1]{\raisebox{0.55ex}{\fontsize{7}{7}\selectfont #1}}
\newcommand{\timestep}{time step}
\newcommand{\tirex}{TiRex}

\newcommand{\moirai}{\textsc{Moirai}}
\newcommand{\moiraimoe}{\textsc{Moirai-MoE}}
\newcommand{\morpheus}{MORPHEUS}
\newcommand{\xlstm}{xLSTM}
\newcommand{\fevbench}{fev-bench}
\newcommand{\tirexemoji}{\raisebox{-0.4ex}{\includegraphics[height=2.5ex]{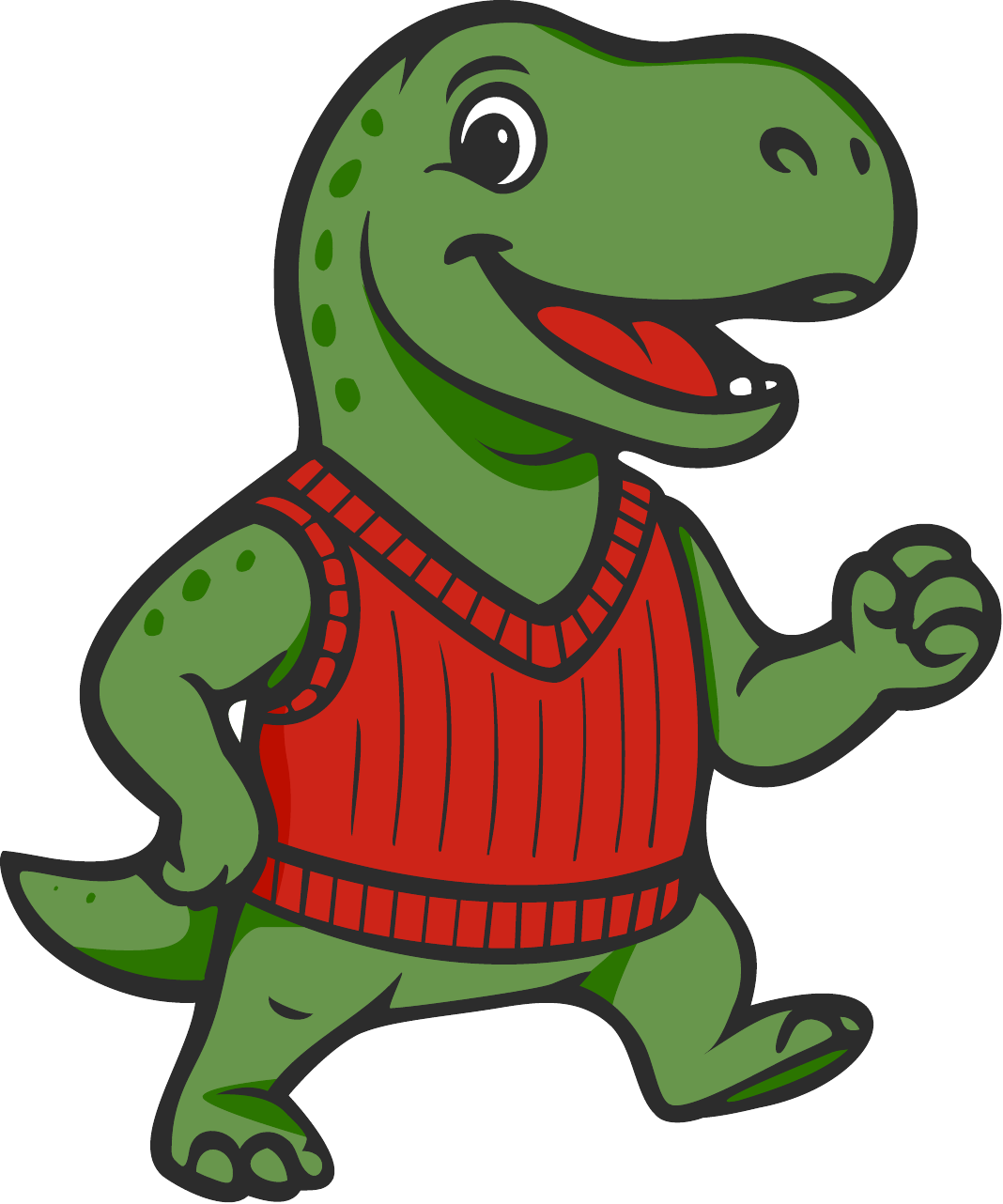}}\hspace{0.2em}}

\newcolumntype{L}[1]{>{\raggedright\arraybackslash}p{#1}}

\nxaisetalogo{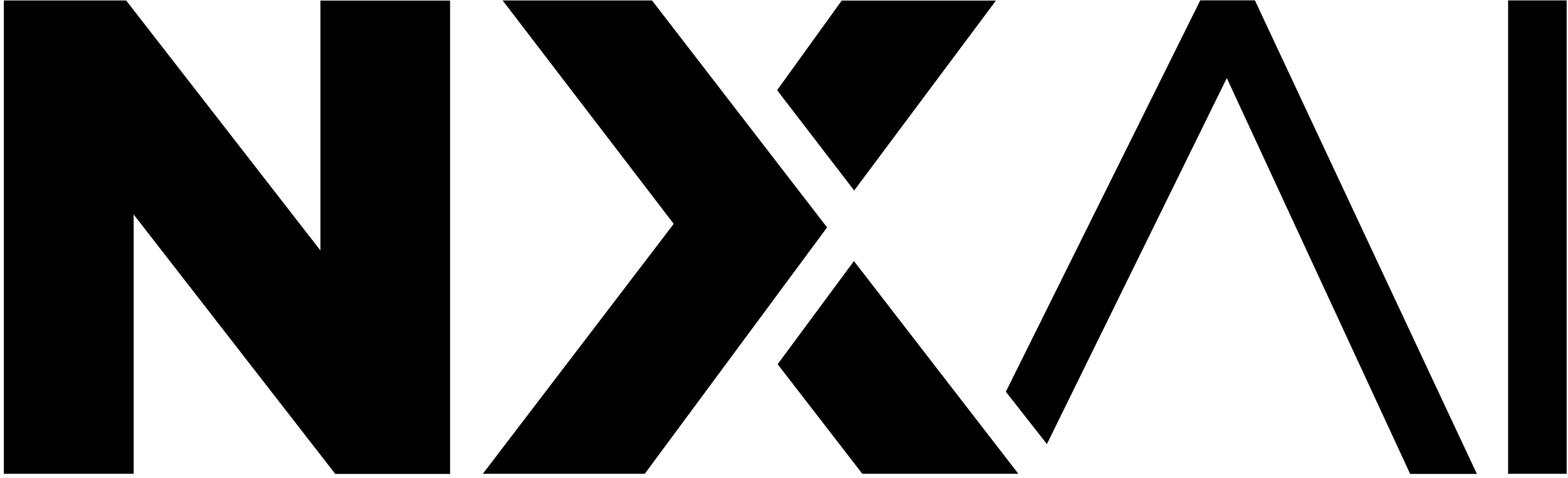}
\nxaisetlogoheight{4mm}
\nxaisetdate{November 23, 2026}

\nxaisetfooter{Accepted as an oral at the 40th Conference on Neural Information Processing Systems (NeurIPS 2026).\\
Corresponding authors: Patrick Podest, \href{mailto:podest@ml.jku.at}{podest@ml.jku.at}, Marco Pichler, \href{mailto:pichler@ml.jku.at}{pichler@ml.jku.at}}

\title{{\tirexemoji TiRex-2: Generalizing TiRex to Multivariate Data and Streaming}}
\author{Patrick Podest\affmark{*,1,2,3}\orcidlink{0009-0003-6546-5706}\and 
Marco Pichler\affmark{*,1,2,3}\orcidlink{0009-0003-4425-6956}\and 
Elias Bürger\affmark{1,2}\orcidlink{0009-0009-6301-6185}\and 
Levente Zólyomi\affmark{1,3}\orcidlink{0009-0008-6392-2471}\and \newline
Bernhard Voggenberger\affmark{3}\orcidlink{0009-0005-0908-9077}\and 
Wilhelm Berghammer\affmark{1,2}\orcidlink{0009-0007-7264-7179}\and 
Daniel Klotz\affmark{4}\orcidlink{0000-0002-9843-6798}\and 
Sebastian Böck\affmark{3}\orcidlink{0000-0001-6707-5427}\and \newline
Günter Klambauer\affmark{1,2,3}\orcidlink{0000-0003-2861-5552}\and 
Sepp Hochreiter\affmark{1,2,3}\orcidlink{0000-0001-7449-2528}
}

\nxaisetaffiliations{%
\affmark{1}ELLIS Unit Linz,  LIT AI Lab \& Institute for Machine Learning, JKU Linz, Austria, \quad
\affmark{2}NXAI Lab, Linz, Austria, \newline
\affmark{3}NXAI GmbH, Linz, Austria, \quad
\affmark{4}Interdisciplinary Transformation University Austria, Linz, Austria, \newline
\affmark{*}Equal contribution.
}

\nxaiabstract{%
We introduce \tirex{-2}, a recurrent xLSTM-based \timeseries{} foundation model that
generalizes the univariate \tirex{} to multivariate forecasting with both past and future
covariates. Real-world forecasting is inherently sequential: observations arrive
continuously, variables evolve jointly, and a subset of covariates is known ahead of time.
Existing Transformer-based time series foundation models capture cross-variate dependencies, but the cost of ingesting each new observation grows at least linearly with context length, even with KV caching.
\tirex{-2} addresses these limitations with a memory-centric recurrent design built around an
asymmetric group-attention variate mixer: paired with a bidirectional time mixer, it
integrates future-known covariates while keeping target variables strictly causal. 
To our knowledge no other time series foundation model does both, and it is what makes streaming at
constant per-patch cost possible.
For scalable multivariate pretraining, we propose a synthetic coupling pipeline that
composes multivariate samples on the fly from large univariate corpora. On GIFT-Eval, \tirex{-2} improves on the zero-shot Chronos-2-Synth on all four statistics,
and on Chronos-2 under a matched corpus. On \fevbench{} Chronos-2 stays ahead on
skill score at $2.2\times$ the mean active parameters.  \tirex{-2} uses 38.4M active
parameters in univariate mode, plus 44.1M for multivariate forecasting. \newline Code is available at \href{https://github.com/NX-AI/tirex-2}{\texttt{github.com/NX-AI/tirex-2}}.
}

\begin{document}
\maketitle

\section{Introduction}

\begin{figure}[!b]
    \centering
    \includegraphics[height=5.5cm, width=1.0\textwidth, keepaspectratio]{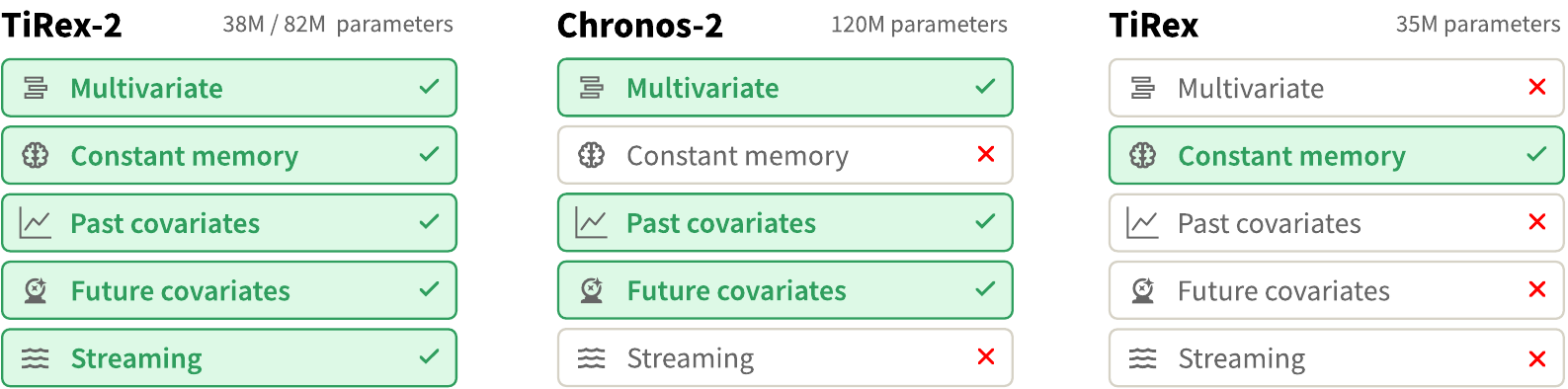}
    \caption{Comparison of \tirex{-2}, Chronos-2, and \tirex{}.
    Chronos-2 supports multivariate forecasting with covariates, but is neither target-causal nor constant-memory. \tirex{-2} adds these properties while preserving native multivariate covariate support (for more details see \Cref{tab:tsfm-comparison}).}
\label{fig:title_comparison}
\end{figure}

Time series arise across diverse domains, including cloud
operations~\citep{joosen_how_2023}, macroeconomics~\citep{sims_macroeconomics_1980}, earth
sciences~\citep{kratzert_rainfallrunoff_2018}, healthcare~\citep{johnson_mimic-iv_2023}, and industrial
monitoring~\citep{sathishkumar_v_e_steel_2021}, where the goal is to extrapolate observed dynamics into the future. Reliable forecasts underpin high-stakes decisions such as flood mitigation~\citep{nearing_global_2024} and predictive maintenance~\citep{yan_comprehensive_2024}. In practice, systems are rarely described by a single signal, rather multiple interacting variates jointly determine the system state. Effective forecasting therefore requires models that capture both temporal structure within each variate and dependencies across variates.

Classical approaches, including autoregressive
models~\citep{hyndman_automatic_2008} and exponential
smoothing~\citep{gardner_exponential_1985}, are typically fit to a single
\timeseries{} and applied to that same series. Vector autoregressive
models~\citep{sims_macroeconomics_1980} extend this paradigm to multiple
variates but remain instance-specific, requiring re-estimation for each
multivariate system. Neural approaches shifted the paradigm toward learning
across collections of related \timeseries{}, with
LSTMs~\citep{Hochreiter:91a,Hochreiter:97} enabling explicit state tracking and
the integration of multiple input variates~\citep{nearing_global_2024}. More
recently, \timeseries{} foundation models (TSFMs) aim to generalize across datasets and domains, with Chronos-2~\citep{ansari_chronos-2_2025}, TimesFM~\citep{das_decoder-only_2024}, \moirai{}~\citep{woo_unified_2024}, and \tirex{}~\citep{auer_tirex_2025} as representative examples.

Despite this progress, a gap remains between univariate scalability and multivariate modelling. Strong univariate foundation models, such as \tirex{}, Reverso~\citep{fu_reverso_2026}, and FlowState~\citep{graf_flowstate_2025}, often rely on a channel-independence assumption when applied to multivariate data, thereby neglecting cross-variate dependencies. Multivariate foundation models, including Chronos-2, \moirai{}, and GTT~\citep{feng_general_2024}, address this limitation but are predominantly based on Transformer architectures. While effective at modeling joint dependencies, these models incur inference costs that grow with context length and require repeated processing of the full history as new observations arrive. This scaling behavior is fundamentally misaligned with streaming forecasting, where predictions must be updated continuously and efficiently.

In this work, we introduce \tirex{-2}, a recurrent \timeseries{} foundation model that extends \tirex{} to multivariate forecasting with both past and future covariates while enabling true streaming inference. The model adopts a memory-centric design based on xLSTM, allowing constant-cost state updates as new data arrive. Architecturally, \tirex{-2} combines a bidirectional time mixer with an asymmetric group-attention variate mixer, enabling the integration of future-known covariates without violating causality over target variables.

\begin{nxaiinfo}[Our contributions are as follows (see also \Cref{fig:title_comparison}):]
\begin{itemize}
\item \textbf{Asymmetric group attention.}
Asymmetric group attention allows future covariates to be used non-causally while keeping targets strictly causal, a first among TSFMs. This extends \tirex{} to jointly model multiple target variates with observed (past) and future covariates.
Additionally, the model bypasses the variate mixer in univariate mode, activating only 38.4M parameters, and it activates the additional 44.1M only in multivariate mode. 

\item \textbf{Streaming inference at constant cost.}
The recurrent state allows incremental forecast updates with constant computation and memory per patch, whereas attention-based models incur costs growing with context length and, if non-causal, require full-context re-computation. Even after 32M streamed \timestep{}s, forecast quality shows no sign of drift.

\item \textbf{Synthetic multivariate coupling for pretraining.}
We introduce a data generation pipeline that constructs diverse multivariate training instances on the fly from univariate corpora, including indirect, causal, and direct cross-variate dependencies, thereby expanding the effective training distribution. Each mechanism contributes measurably.
\end{itemize}
\end{nxaiinfo}

\Cref{sec:related-work} reviews related work, \Cref{sec:model} presents the
setting, architecture, and coupling pipeline, and \Cref{sec:experiments}
reports the evaluation.

\section{Related work}
\label{sec:related-work}
\paragraph{Univariate TSFMs.}
Early transformer-based TSFMs are either 
pretrained LLMs \citep{xue_promptcast_2023, gruver_large_2023} or trained on time-series data directly
\citep{ansari_chronos_2024}.
These approaches commonly pass each \timestep{} individually to the model.
However, splitting \timeseries{} into non-overlapping patches \citep{nie_time_2022} has become the dominant architectural choice for TSFMs \citep{liu_sundial_2025, das_decoder-only_2024, wang_output_2025, liu_moirai_2026}. 
Recent work has explored large mixture of experts models \citep{liu_timer-s1_2026, wu_wavemoe_2026}
with billions of parameters, \citet{liu_moirai-moe_2025} suggesting that the different experts specialize based on the shape of patches.
In contrast, smaller recurrent neural network (RNN) approaches remain
competitive, both without \citep{fu_reverso_2026, graf_flowstate_2025} and with \citep{auer_tirex_2025} patching.
In this work, we adapt the recurrent \tirex{} architecture proposed by \citet{auer_tirex_2025} and extend it via the
variate-mixing attention layers to learn cross-correlation patterns. 

\paragraph{Multivariate TSFMs.} Two architectural strands dominate joint
multivariate forecasting. The first flattens all variates into one sequence with
any-variate attention: \moirai{}~\citep{woo_unified_2024} and
\moiraimoe{}~\citep{liu_moirai-moe_2025} concatenate all variates into a single
sequence, while \morpheus{} \citep{patil_morpheus_2025} interleaves individual
timesteps with separation tokens in between. As single-target variants,
COSMIC~\citep{auer_zero-shot_2025} and TimesFM-ICF~\citep{faw_-context_2025}
use the other variates to enhance the
prediction (the covariate setting). Flattening, however, inflates sequence length and limits per-variate
context. The second strand factorizes time and variate attention into separate layers,
introduced by Crossformer \citep{zhang_crossformer_2022} and adopted for TSFMs
by \citet{feng_general_2024} and Chronos-2 \citep{ansari_chronos-2_2025} to
scale to $8{\rm k}$ \timestep{}s and many variates; \citet{liu_itransformer_2023} push
this further by embedding entire \timeseries{} into single tokens.
 
Models differ further in whether they accept \emph{past-only} or
\emph{future-known} covariates (whose future values are available at inference),
and those handling both alongside multivariate targets remain
rare~(\Cref{tab:tsfm-comparison}). A complementary line instead adapts
univariate TSFMs through covariate-aware projections, decomposition, trainable
variate-mixing layers, or in-context regression on
covariates~\citep{benechehab_adapts_2025, arango_chronosx_2025, cheng_cora_2026,
ekambaram_tiny_2024, chen_tsmixer_2023, auer_zero-shot_2025, burger2026covariates}. We adopt
factorized time/variate
attention~\citep{gao2024units,cohen_this_2025,zhang_crossformer_2022} on top of
the recurrent \tirex{} backbone, yielding memory \emph{linear} in sequence
length while supporting both covariate types with multivariate targets, allowing
the model to leverage, e.g., historical sensor readings alongside future-known
calendar or promotion features, and adaptively disable variate-mixing for
univariate inputs to preserve the efficiency of \tirex{}.  
 
\paragraph{Synthetic data generation.}
   In data-scarce settings, synthetic data is a viable option for training
   TSFMs~\citep{ansari_chronos-2_2025,oreshkin_zero-shot_2026,moroshan2026tempopfn}.
   Existing generators mostly target univariate \timeseries{}, using adapted
   seasonal ARIMA processes~\citep{oreshkin_zero-shot_2026}, Gaussian
   processes~\citep{ansari_chronos_2024}, or temporal causal models
   (TCMs)~\citep{xie_cauker_2025,runge_causal_2023}. For multivariate
   \timeseries{}, \citet{ansari_chronos-2_2025} mention unspecified
   ``multivariatizers'' that couple independently sampled univariate series. We
   make this step explicit with a concrete, largely TCM-based framework of coupling
   mechanisms for generating diverse synthetic multivariate dependencies.

\section{A TiRex architecture for multivariate forecasting with covariates}
\label{sec:model}
\subsection{Problem setup}

In multivariate \timeseries{} forecasting,
we want to forecast $F_x$ target series $\BX = [\Bx_1,\ldots,\Bx_{F_x}]^\top$, each
spanning $T$ \timestep{s} of historical context and a prediction horizon of $O$
\timestep{s}, i.e. $\Bx_{i,1:T}\in\reals^{T}$ and $\Bx_{i,T+1:T+O}\in\reals^{O}$.
If available, the model is additionally conditioned on $F_y$
past covariates $\BY\in\reals^{F_y\times T}$ and $F_z$ future-known covariates $\BZ\in\reals^{F_z\times (T
+ O)}$, known across the entire prediction horizon (e.g., calendar features or scheduled interventions). Then the goal is to model the conditional distribution
\begin{equation*}
  \cP \bigl( \BX_{T+1:T+O} \bigm| \BX_{1:T}, \BY_{1:T}, \BZ_{1:T+O} \bigr).
\end{equation*}

To this end, the model approximates the distribution above using $Q$ evenly
spaced quantiles, conditioned on the available context: $\hat{\BX}_{T+1:T+O} = [\hat{\BX}_{k,T+1:T+O}]_{k=1}^{Q}$, where $\hat{\BX}_{k,T+1:T+O}\in\reals^{F_x\times O}$ denotes the $k^\text{th}$ quantile prediction.\enlargethispage{\baselineskip}

\subsection{Architecture}

\newlength{\panelH}
\setlength{\panelH}{8.6cm}  
\begin{figure}[ht]
    \centering
    \includegraphics[width=\textwidth]{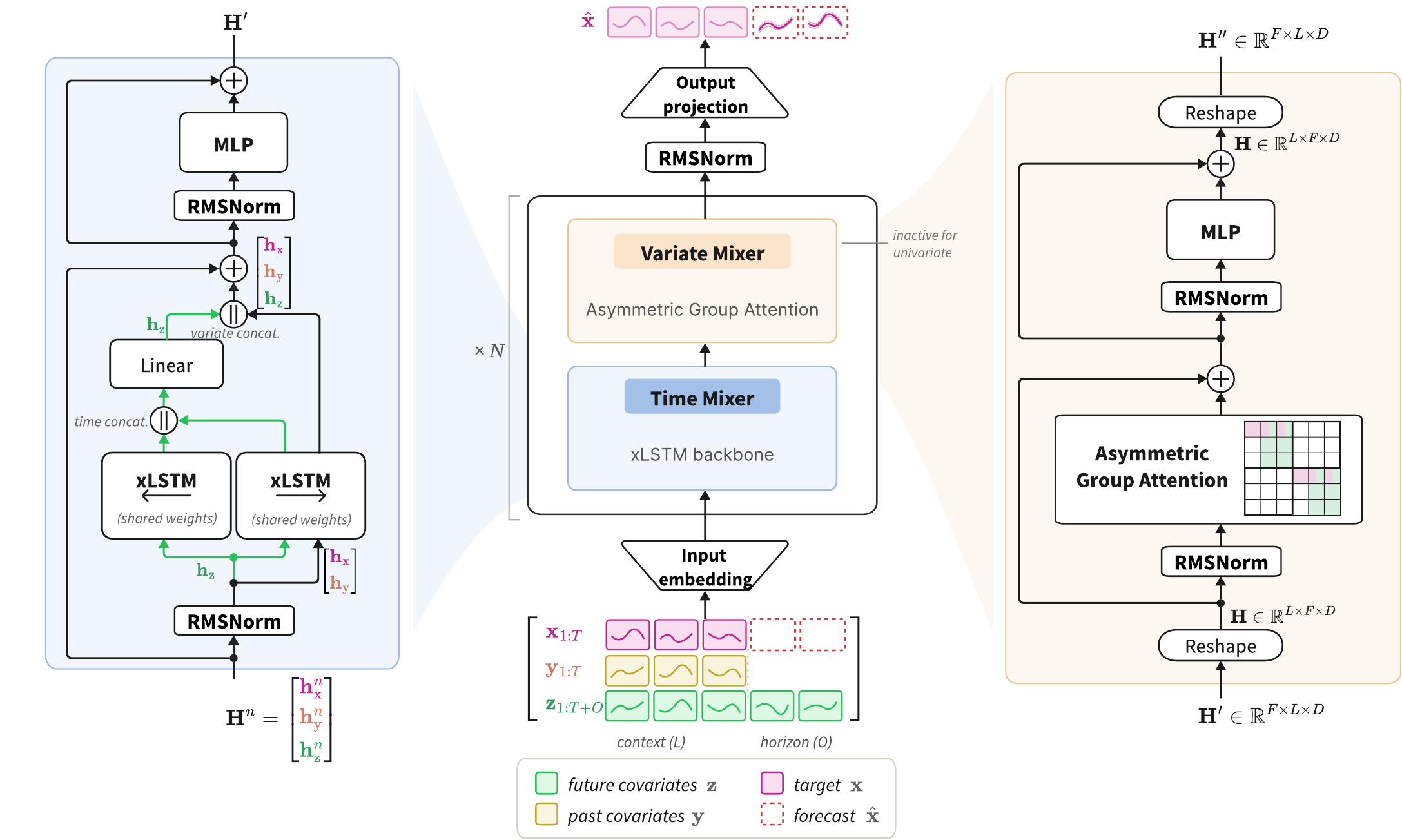}\\[-0.2em]
    \begin{subcaptionblock}{0.3\linewidth}\caption{Time mixer}\label{fig:arch:time}\end{subcaptionblock}
    \begin{subcaptionblock}{0.3\linewidth}\caption{Overall
    architecture}\label{fig:arch:overview}\end{subcaptionblock}
    \begin{subcaptionblock}{0.3\linewidth}\caption{Variate
    mixer}\label{fig:arch:variate}\end{subcaptionblock}
    \caption{
      \tirex{-2} alternates time- and variate-mixing blocks. The variate mixer's
      asymmetric group attention is what allows future-known covariates to be
      exploited bidirectionally without leaking future-known targets into earlier
      positions. (\subref{fig:arch:overview}) Each multivariate \timeseries{} is
      split into non-overlapping patches before being embedded into tokens. The
      stack mixes information across time (time mixer), across variates (variate
      mixer) and inside each token (MLP). The output-projection produces $Q$ quantile
      forecasts per \timestep{} in each output patch. (\subref{fig:arch:time})
      The time mixer applies a forward xLSTM to all variates, plus a weight-tied
      reverse pass for future-known covariates, fused linearly.
      (\subref{fig:arch:variate}) The variate mixer applies group attention
      along the variate axis with an asymmetric mask preventing target-to-covariate flow.
      Note: we omitted the layer super-scripts of the token
      representations $\BH$ inside the blocks
      for clarity. 
   }
    \label{fig:arch}
\end{figure}

\paragraph{Block signature.}
\Cref{fig:arch:overview} shows the high-level architecture of \tirex{-2}. The model first pads targets and past covariates to
length $T+O$, then splits every variate independently into $L=\lceil(T+O)/P\rceil$ non-overlapping patches of size $P$
and embeds each patch into a token, yielding a token tensor $\BH\in\reals^{F\times L\times D}$, with $F = F_x + F_y + F_z$
the total number of variates and $D$ the token embedding dimension.
\tirex{-2} then alternates a time mixer, which mixes information across the $L$ axis independently per variate, with a
variate mixer, which mixes across the $F$ axis independently per patch.

\paragraph{Input layer.}
The input layer maps each raw variate $\Bu\in\reals^{T+O}$, which is a row of either $\BX_{1:T}$, $\BY_{1:T}$ or
$\BZ_{1:T+O}$ padded to length $T+O$, to a sequence of $L$ tokens in three
steps: scaling, patching, and embedding.
Scaling builds on reversible instance normalization~\citep{kim_reversible_2021} to mitigate distribution
shift across variates, and adds the inverse hyperbolic sine transform
from econometrics~\citep{burbidge_alternative_1988} and energy price forecasting~\citep{uniejewski_efficient_2018}. This prevents 
heavy-tailed variates from dominating the loss.
Concretely, we standardize $\Bu$ with its empirical mean $\hat{\mu}$ and standard deviation
$\hat{\sigma}$ computed over the observed context ($t<T$, ignoring missing entries) and apply
\begin{equation*}
  \tilde{u}_t \;=\; (1-b)\operatorname{arcsinh}\!\PAR{\frac{u_t-\hat{\mu}}{\hat{\sigma}}} + b\,u_t,
\end{equation*}
which behaves linearly near the origin and logarithmically in the tails. The binary gate
$b$ handles a separate pathology: sparse binary covariates are degenerate under naive standardization,
since the gap between the two standardized levels is $1/\sqrt{p(1-p)}$, which
scales as $1/\sqrt{p}$ for rare positive classes ($p\to 0$) and
diverges symmetrically as $p\to 1$. We therefore detect binary variates
with the indicator $b=\BOn[\forall\, 0\!\leq\!t\!<\!T: x_t\in\{0,1\}]$ and
bypass the affine transform when $b=1$, preserving the canonical $\{0,1\}$
encoding regardless of context-window sparsity. We defer the full specification to
\Cref{app:scaler}.
After scaling, each variate is split into $L=\lceil(T+O)/P\rceil$ non-overlapping patches of length $P$; 
unobserved positions (future targets, future past-covariate values, and missing entries) are padded.
A two-layer residual MLP shared across variates and patches $\reals^{P}\!\to\!\reals^{D}$ then embeds each patch into a token of dimension $D$, yielding the
input tensor $\BH$ consumed by the first block.

\paragraph{Time mixer.}
The time mixer acts along the patch axis $L$ independently for every variate, with directionality determined by
the variate's type. Target and past-covariate tokens are processed by a strictly forward
\xlstm{}~\citep{beck_xlstm_2024}. Future-known covariate tokens are additionally processed in reverse by the
\emph{same} weight-tied \xlstm{} \citep{schmidinger2025bioxlstm}, and the two directions are combined by a linear fusion layer, so that each
future-known covariate token encodes information from both before and after its position. Concretely, the time mixer maps
the input tensor $\BH$ to $\BH'$. Writing 
$\bar{\BH}=\mathrm{RMSNorm}(\BH)$ for the pre-normalized input, $1\leq l \leq L$ indexing the patches, $\theta$ the
(weight-tied) parameters of the \xlstm{}, and
$[\,\cdot\,,\cdot\,]$ and $[\,\cdot\,;\cdot\,]$ for concatenation along the sequence and variate axes, respectively,
\begin{alignat*}{2}
  \tilde{\Bh}_{x,l} &= \text{\xlstm{}}\!\PAR{\bar{\BH}_{x,1:l}; \theta}, &\qquad
    \overrightarrow{\Bh}_{z,l} &= \text{\xlstm{}}\!\PAR{\bar{\BH}_{z,1:l}; \theta}, \\
  \tilde{\Bh}_{y,l} &= \text{\xlstm{}}\!\PAR{\bar{\BH}_{y,1:l}; \theta}, &\qquad
    \overleftarrow{\Bh}_{z,l} &= \text{\xlstm{}}\!\PAR{\bar{\BH}_{z,L:l}; \theta}, \\
  \tilde{\Bh}_{z,l} &= \mathrm{Linear}\!\PAR{\SBRA{\overleftarrow{\Bh}_{z,l}, \overrightarrow{\Bh}_{z,l}}}, &\qquad
    \tilde{\Bh}_l\hphantom{{}_{z,}} &= \SBRA{\,\tilde{\Bh}_{x,l}\,;\, \tilde{\Bh}_{y,l}\,;\, \tilde{\Bh}_{z,l}\,} \in \reals^{F\times D}.
\end{alignat*}
The time-mixed output $\BH' = \BH + \tilde{\BH}$, where
$\tilde{\BH} = [\tilde{\Bh}_l]_{l=1}^{L}$ further passes through a residually connected MLP creating the final block
output.
The time mixer processes just the future known covariates twice (forward and reverse) while targets and past covariates remain strictly forward, making streaming forecasts well-defined (Sec.~\ref{subsec:streaming}).

Following \citet{beck_xlstm_2024}, who report benefits from mLSTM blocks on
   long-context tasks, we instantiate the time mixer as an \xlstm{} stack that
   alternates mLSTM and sLSTM blocks, rather than using the sLSTM-only backbone of
   the original \tirex{}~\citep{auer_tirex_2025}.

\paragraph{Variate mixer.}
The number of variates $F$ varies from series to series and there is no
canonical ordering. We therefore implement the variate mixer as multi-head
self-attention, which natively handles arbitrary $F$ and is
permutation-equivariant within each variate type.
We use block-diagonal \emph{group}
attention~\citep{feng_general_2024,cohen_this_2025,ansari_chronos-2_2025}, which
prevents interactions across concatenated series and avoids padding to the
largest $F$ in a batch.
Within each group we further apply an \emph{asymmetric mask} $M$: target queries may read covariate keys, but covariate
queries cannot read target keys. With $1\leq i,j \leq F$ indexing (query,\,key) within the group and $\cX,\cY,\cZ$ the target, past- and future-covariate index sets, we have:
\begin{equation}
  m_{ij} =
  \begin{cases}
    0 & \text{ $\neg(i\!\in\!\cY\cup\cZ\wedge j\!\in\!\cX)$,}\\
    -\infty & \text{otherwise.}
  \end{cases}\label{eq:variate-mask}
\end{equation}

Entries $m_{ij}=-\infty$ pull the attention weight to 0 in the
$\soft$ operation preventing information flowing from token
$j\rightarrow i$ (target to covariate) while entries $m_{ij}=0$ allow it (covariate to
target). 

\begin{lemma}[One-block target dependence]\label{lem:one-block}
  Let $\Bh^{n}_{x,l}$ denote the $l^{th}$ target token after the first $n$ blocks. 
  Then $\Bh^{n+1}_{x,l}$ is a function only of
  $\BH^{n}_{x,1:l}$, $\BH^{n}_{y,1:l}$, and 
  $\BH^{n}_{z,1:L}$.
\end{lemma}
The forward-only \xlstm{} restricts target and past-covariate tokens at patch
$l$ to indices $\leq l$, while the asymmetric mask in~\Cref{eq:variate-mask}
blocks every covariate query from reading a target key, so the variate mixer
cannot reintroduce a future-known target dependency through the covariate channel. 

\begin{proposition}[Target-causality]
  \label{prop:target-causality}
  By induction on the block depth using \Cref{lem:one-block}, for every $n\leq N$
  and patch index $l$ the target token $\Bh^{n}_{x,l}$ does
  not depend on $\Bh^{0}_{x,l'}$ for any $l'>l$.
\end{proposition}
Intuitively, future-known covariates may carry information backward along the patch
axis, but the mask prevents those tokens from ever reading target tokens.
Combined with the output head, the next-patch prediction emitted from
position $l$ therefore cannot read the target patch it is trained to predict. 
A full proof is given in \Cref{app:proof-of-causality}.

To our knowledge, this makes \tirex{-2} the first TSFM that exploits
future-known
covariates bidirectionally while keeping target streams strictly causal in a
single forward pass. The closest prior designs  are TimeXer~\citep{wang_timexer_2024},
Timer\,XL~\citep{liu_timer-xl_2024} and CITRAS~\citep{yamaguchi_citras_2025}. All of them
impose related but weaker asymmetries and are task-specific or fully causal
along time. \Cref{app:prior-cross-variate} provides a detailed comparison.

\paragraph{Output layer and loss.} A residual, two-layer MLP maps the target
hidden representations
$\BH_x\in\reals^{F_x\times L\times D}$ produced by
the final time and variate mixers to patch-wise next-patch quantile forecasts
\citep[following][]{auer_tirex_2025},
$\hat{\BX}\in\reals^{F_x\times Q\times (L-1)\times
P}$, i.e., $Q$ quantiles for each of the $P$ \timestep{s} of every output patch.
Predictions are returned to the original data domain by applying the inverse of the input scaler, which clips its argument to a conservative bound before applying $\sinh$ to suppress implausibly large outlier predictions and prevent overflow in the output datatype (see \Cref{app:scaler}).
The model is trained with the pinball loss~\citep{koenker_regression_1978}
applied at every output \timestep{} (not only on the horizon), giving up to an $(L{-}1)$-fold denser gradient signal than supervising on the horizon alone:
\begin{equation*}
 \cL = \frac{1}{F_x\,Q\,|\cT_\text{obs}|}
    \sum_{t\in\cT_\text{obs}} \sum_{f=1}^{F_x} \sum_{k=1}^Q
    \big[\,q_k\,\max(x_{f,t}-\hat{x}_{f,k,t},\,0) + (1-q_k)\,\max(\hat{x}_{f,k,t}-x_{f,t},\,0)\,\big],
\end{equation*}

where $\cT_\text{obs}$ is the set of observed (non-missing) target \timestep{s} and $q_k$ denotes the
$k^{\text{th}}$ quantile level. Missing values are excluded from both sum and normalisation. 

\subsection{Long-context efficiency and streaming forecasts}
\label{subsec:streaming}
The recurrent \xlstm{} time mixer keeps the per-block forward time complexity at
$\cO(L)$ rather than the $\cO(L^{2})$ of attention, the variate dimension $F$ stays constant and therefore does not affect scaling in $L$. The maintained hidden state of \xlstm{} additionally enables constant-time updates. Streaming
workloads predominantly involve target and past-covariate streams observed up to
the current \timestep{}, both routed through the forward-only \xlstm{}. In an
online forecasting setting, which we refer to as streaming, each newly arrived
patch can be ingested and a forecast emitted in constant time. In contrast, a
transformer-based time mixer using KV-caching would pay $\cO(L)$ per new
patch, its latency growing linearly with the number of patches $L$.
\tirex{-2} can therefore be fed continuously and emit forecasts in time
proportional to the length of the increment, not to the full lookback horizon.

\subsection{Synthetic multivariate coupling}
\label{sec:uni2multi}

\begin{figure}[!ht]
    \centering
    \includegraphics[height=6cm, width=1.0\textwidth, keepaspectratio]{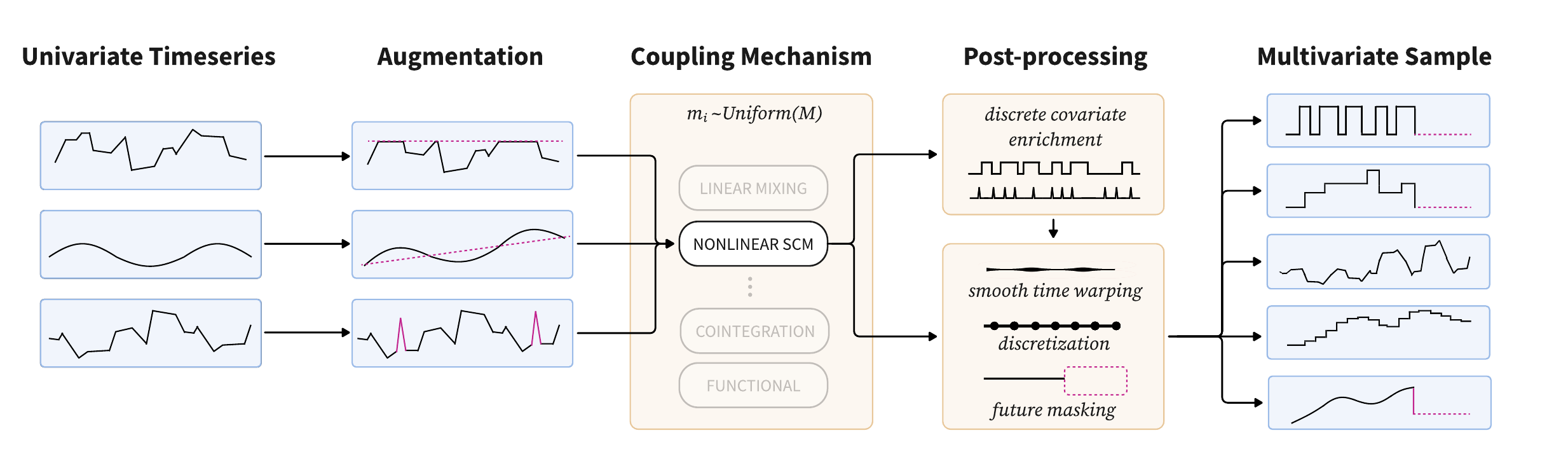}
    \caption{Synthetic multivariate coupling pipeline. A batch of univariate series is first independently augmented (amplitude trends, censoring, spike injection). A coupling mechanism $c_i \sim \mathrm{Uniform}(C)$ is then sampled from $C =\{$ identity, univariate, linear mixing, linear structural causal model (SCM), nonlinear SCM, cointegration, functional $\}$ and, except for the identity and univariate pass-through cases, transforms the augmented series into jointly dependent variates. Post-processing adds realistic structure via covariate enrichment and applies smooth time warping, discretization, and future masking, producing the final multivariate training sample.}
    \label{fig:pipeline}
\end{figure}

A model that natively consumes multivariate inputs with covariates only learns to use them if the training distribution actually contains diverse cross-variate dependencies. Existing curated multivariate corpora are too narrow to enforce this learning process, while large univariate corpora are abundant. We bridge this gap with a synthetic coupling pipeline that, at training time, draws a batch of univariate series from a shared pool and \emph{couples} them on the fly into multivariate samples with controllable cross-variate dependencies (Fig.~\ref{fig:pipeline}):

Each series is first independently perturbed with piecewise-linear amplitude trends, quantile censoring, and synthetic spikes (Gaussian, triangular, or rectangular kernels), then cropped or NaN-padded to
length~$T$. Given $F$ such augmented series $\Bu_1,\dots,\Bu_F\!\in\!\reals^{T}$, one of the following mechanisms is sampled to produce $F$ output variates $\Bx_1,\dots,\Bx_F\!\in\!\reals^{T}$, written entrywise as $x_{f,t}$, with a known dependency structure:

\newpage
\begin{enumerate}
  \item \textbf{Identity / pass-through:} $x_{f,t}=u_{f,t}$ or a single univariate output, preserving univariate forecasting ability as a no-coupling control.

  \item \textbf{Functional coupling:} $x_{f,t}=\phi_f(u_{0,t})+\varepsilon_{f,t}$ with monotone, compressive, discretizing, or piecewise-linear $\phi_f$, yielding direct pointwise dependence as in sensor redundancies.

  \item \textbf{Linear mixing:} $x_{f,t}=\sum_{i=1}^{F} a_{fi}\,u_{i,t}$, with the singular-value spectrum of the mixing matrix $A=(a_{fi})$ sampled from dominant, uniform, or power-law regimes, mimicking shared latent drivers as in factor models.

  \item \textbf{Cointegration:} $x_{f,t}=\sum_{k} \Lambda_{fk}\,s_{k,t}+\xi_{f,t}$ with shared random-walk trends $s_{k,t}$ and stationary AR(1) residuals $\xi_{f,t}$, reproducing long-run equilibria between nonstationary variates.

  \item \textbf{Linear structural causal model:} a random directed acyclic graph with lagged edges, \newline $x_{f,t}=\sum_{i\in\mathrm{pa}(f)}\alpha_{if}\,u_{i,\,t-\tau_{if}}+\varepsilon_{f,t}$, introducing directed lead--lag structure.

  \item \textbf{Nonlinear structural causal model:} using nonlinearities $g_{if}$ and an optional multiplicative gate $h$, $x_{f,t}=h(u_{k,\,t-\tau_k})\sum_{i\in\mathrm{pa}(f)} g_{if}(u_{i,\,t-\tau_{if}})$, adding state-dependent coupling.
\end{enumerate}
The resulting samples are finally enriched with realistic covariate structure: variate permutation, smooth
per-variate time warping via Brownian-bridge lags, patch masking with contiguous NaN blocks \citep{auer_tirex_2025} per-variate, partial future observability by
truncating future portions of random covariates, and discretization in
value (uniform, quantile, power-law) and time (freezes, staircases,
duty cycles). See \Cref{app:coupler} for details.

\section{Experiments}
\label{sec:experiments}
We evaluate \tirex{-2} along two axes. First, we compare against other zero-shot \timeseries{} foundation models on two separate benchmarks, \fevbench{}~\citep{shchur_fev-bench_2025} and GIFT-Eval~\citep{aksu_gift-eval_2024}. Second, we use synthetic data to isolate streaming behavior, long-horizon forecasting, and covariate-shift sensitivity. We conclude with architectural ablations that quantify the contribution of the main components of \tirex{-2}.

\paragraph{Training setup.}
We pretrain \tirex{-2} for $700{,}000$ steps on $2$ NVIDIA H100 GPUs, followed by a short long-context post-training phase at context length of $8{,}192$ and prediction horizon $512$ \timestep{s} to adapt the model to longer sequences. The backbone consists of $N = 12$ alternating \mbox{mLSTM}/sLSTM \xlstm{} blocks. Full hyperparameters and training details for both phases are given in \Cref{app:training-setup}.

\paragraph{Evaluation models.}

In the following experiments, we compare \tirex{-2} to publicly available
\timeseries{} foundation models on \fevbench{}~\citep{shchur_fev-bench_2025} and
GIFT-Eval~\citep{aksu_gift-eval_2024}, restricting to models for which
zero-shot evaluations and inference code were published on the respective
leaderboard. This yields two benchmark-specific comparison sets: on \fevbench{},
Chronos-Bolt, \moirai{-2}, Toto-1.0~\citep{cohen_toto_2024},
\tirex{}~\citep{auer_tirex_2025}, TimesFM-2.5~\citep{das_decoder-only_2024}, and
Chronos-2~\citep{ansari_chronos-2_2025}; \Cref{fig:fev_bench_scores} additionally
includes TabPFN-TS~\citep{hoo_tables_2026}, Sundial-Base~\citep{liu_sundial_2025}
and the statistical baselines of the benchmark itself. On GIFT-Eval, Chronos-2-Synth, PatchTST-FM-r1 (Granite and
base)~\citep{wen_revisiting_2026},
\tirex{}, TimesFM-2.5, and
FlowState-r1.1~\citep{graf_flowstate_2025}.
On the other experiments we compare \tirex{-2} against purely multivariate TSFMs
on synthetic data, and therefore relax the zero-shot constraint. We include
Moirai (1.0 and MoE)~\citep{woo_unified_2024,liu_moirai-moe_2025},
GTT~\citep{feng_general_2024}, Toto, and Chronos-2.

\subsection{Zero Shot}
\label{sec:experiments:zeroshot}

\begin{figure}[!ht]
  \vspace{-0.3cm}
  \centering
  \includegraphics[width=\linewidth]{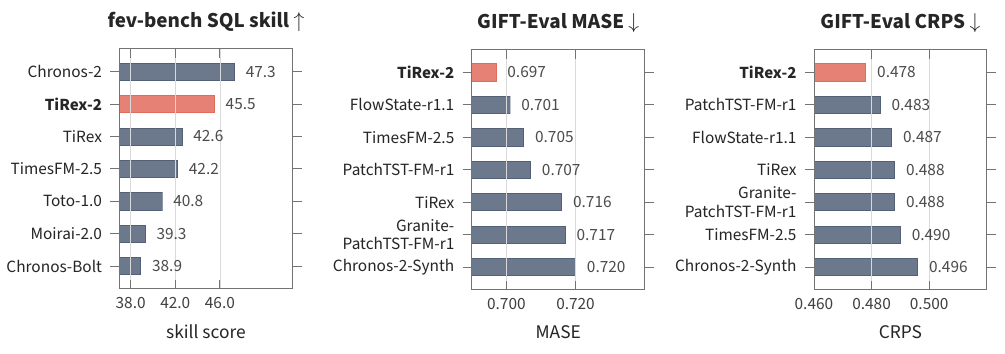}
\caption{Zero-shot performance of \tirex{-2} against the best
\timeseries{} foundation models on each metric: the \fevbench{} SQL skill
score against Seasonal Naive, where higher is better, and GIFT-Eval MASE and
CRPS, where lower is better. Panels are sorted with the best model at the top,
so the leaderboard position is read the same way in all three.}
\label{fig:benchmark_comparison}
\end{figure}

\begin{figure}[!b]
  \centering
  \includegraphics[width=\linewidth]{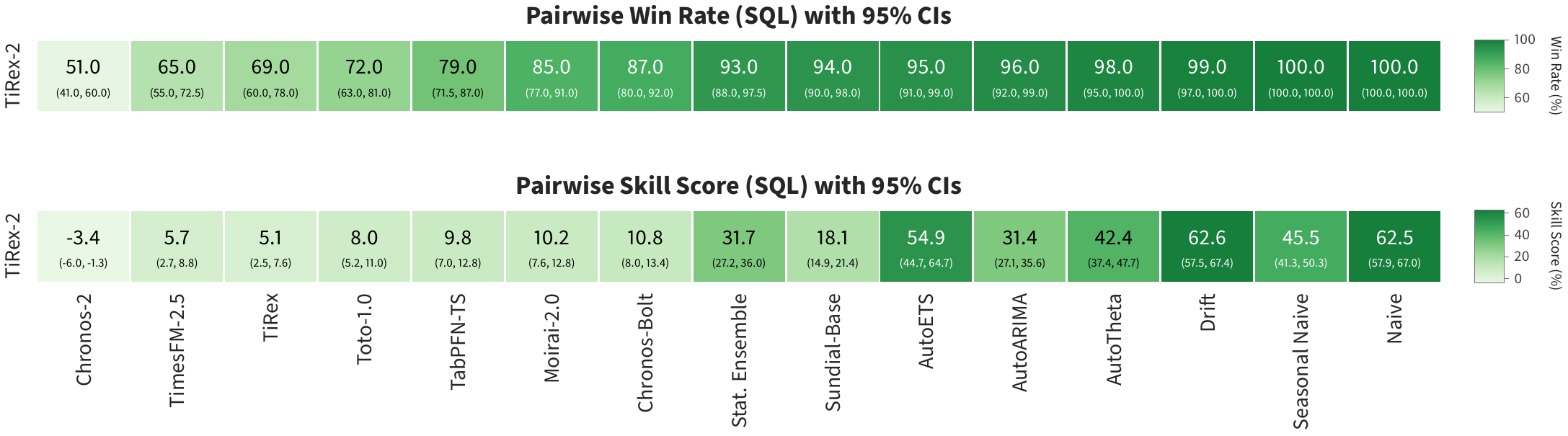}
\caption{Zero-shot performance of \tirex{-2} against \timeseries{}  models on \fevbench{} (SQL) using both Pairwise Win Rate and Pairwise Skill Score with 95\% confidence intervals.}
\label{fig:fev_bench_scores}
\end{figure}

We evaluate zero-shot forecasting on two complementary benchmarks. \fevbench{}~\citep{shchur_fev-bench_2025}
probes the ability to exploit past and future-known covariates, whereas GIFT-Eval~\citep{aksu_gift-eval_2024} probes generalization across diverse domains, sampling frequencies, and horizons. To rule out training-test leakage, we pretrain a
separate checkpoint per benchmark, with overlapping datasets removed from the respective
training corpus (see \Cref{app:pretrain-corpus}). Chronos-2 includes train-splits of GIFT-Eval during pretraining, so the zero-shot comparison there is against Chronos-2-Synth, which
\tirex{-2} improves on across all four statistics. Under a corpus matched to Chronos-2,
\tirex{-2} improves on Chronos-2 on all four as well (\Cref{app:add:gift-matched}).
On \fevbench{} Chronos-2 remains ahead (\Cref{fig:benchmark_comparison},
\Cref{app:add:fev-protocol}). \Cref{fig:fev_bench_scores} gives the pairwise picture against every other model on the
leaderboard: \tirex{-2} wins 65--100\% of tasks against all of them but Chronos-2, with
which it splits tasks evenly ($51.0$, CI $41.0$--$60.0$) while trailing by $1.8$ SQL
skill points.

\subsection{Sensitivity to streaming, covariates and forecast horizon}
\label{sec:shift_analysis}

\begin{figure}[!ht]
  \vspace{-2mm}
  \centering
  \includegraphics[width=\linewidth]{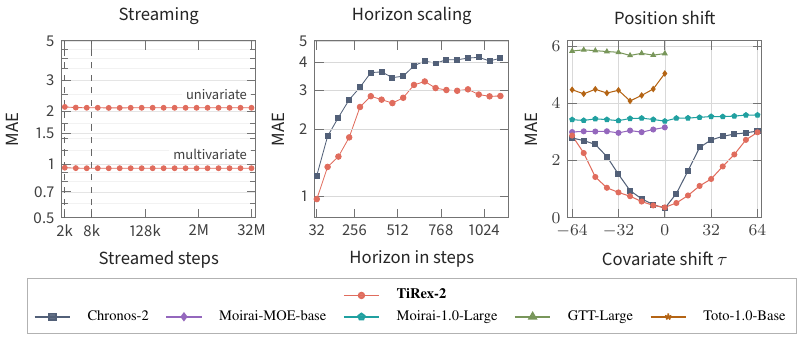}
\caption{%
Left: streaming. MAE is stable far beyond the $2{\rm k}$ pretraining and
$8{\rm k}$ post-training context boundaries (dashed).
Middle: long-horizon forecasting on \texttt{dysts} (log $y$-axis), averaged
over temporal granularities. \tirex{-2} attains lower MAE than Chronos-2 at
all horizons, with the gap widening as the horizon grows (per-granularity
results in \Cref{fig:longhorizon_appendix}).
Right: covariate shift sensitivity. \tirex{-2} retains useful covariate signal
over a wider lag range than Chronos-2.
Setup in Sections~\ref{sec:shift_analysis} and~\ref{app:add:setup}.}
\label{fig:horizon_scaling_and_shift}
\end{figure}

\paragraph{Streaming.}
The causal, recurrent design of \tirex{-2}
(Sec.~\ref{subsec:streaming}) lets the model ingest arbitrarily long contexts
patch by patch at constant per-patch cost for target and past-covariate
streams and emit a forecast after every update. We stream up to $32$M time
steps of a first-order autoregressive target driven by a noisy, lagged
covariate registered as future-known, and report MAE per emitted patch
(Fig.~\ref{fig:horizon_scaling_and_shift}, left). Forecast quality remains
stable far past the $8{\rm k}$ post-training context: the recurrent state
extrapolates cleanly to context lengths $4000\times$ beyond training, without
any sign of saturation or drift. On real data, \citet{pichler2026time} confirm
that \tirex{-2} retains its accuracy on the full history of the ten fev-bench
tasks exceeding $50{\rm k}$ steps, while all other evaluated TSFMs degrade
beyond their trained context. \Cref{app:add:latency} presents additional
results on retained forecast quality as well as latency and memory
measurements.

\paragraph{Long-horizon forecasting.}
Long-horizon forecasting on chaotic systems is effectively a
covariate-utilisation stress test: models must extract signal from the remaining state variables to maintain accuracy at long horizons. The
\texttt{dysts}~benchmark
\citep{gilpin_chaos_2023} provides 135
chaotic trajectories at three temporal granularities. We forecast one
channel as target with the rest as future-known covariates, evaluating \tirex{-2}
and Chronos-2 at $h\!\in\!\{32, 64, \ldots, 1056\}$.

\paragraph{Covariate shift sensitivity.}
Since real-world covariates are rarely perfectly aligned with the target
\citep[e.g.][]{podobnik_time-lag_2010,zhao_investigating_2023}, we probe
shift tolerance by pairing synthetic random-walk targets with a
shifted, noisy covariate, $z_t^\tau = x_{t-\tau} +
\varepsilon_t$, $\varepsilon_t\!\sim\!\cN(0,\,0.1^2)$, where $x$ is the target and $\tau$ the shift in
\timestep{s}, and tracking
the MAE against $\tau$
(Fig.~\ref{fig:horizon_scaling_and_shift}). \tirex{-2} remains informative
well beyond the range where Chronos-2 has fallen back to its no-covariate
baseline. 
For Toto-1.0, GTT and Moirai-MOE the covariate was provided as a past covariate
since this is the setting which the models support, one can therefore see that
the models only make gains for negative shifts, where the future is shifted into the past.

\subsection{Ablations}
\label{sec:ablations}
We ablate the core design choices of \tirex{-2} on \fevbench{},
which contains enough genuinely multivariate tasks to produce
meaningful signal. Moreover, we investigate the impact of each mechanism of the synthetic multivariate coupling on \fevbench{} and compare those impacts on GIFT-Eval. Each configuration is trained with up to six seeds.
We report mean and standard deviation of the SQL and MASE skill score metrics for \fevbench{}, and the MASE and CRPS metrics for GIFT-Eval.

For the architecture ablations, relative to the full model, we remove or
replace group attention, asymmetric attention masking, binary-aware scaling, bidirectional context
mixing, and the mixed sLSTM/mLSTM backbone. We further test the
pre-processor and future-known covariates. Providing future-known covariates
yields a consistent gain, indicating that forecast-window signal
outweighs extended autoregressive history (\Cref{tab:ablations}).

In the coupling mechanism ablation, we disable each coupler separately and show
the results in \Cref{tab:coupler-ablation}. In the multivariate evaluation
(\fevbench{}), removing any single coupler lowers both skill scores by roughly
2--3 points, so each mechanism contributes; the nonlinear SCM causes the largest
drop in SQL, while identity and functional coupling matter most for MASE.
As expected, on the largely univariate GIFT-Eval the effects are an order of
magnitude smaller and mostly within seed variance, with functional coupling
the only coupler whose removal measurably hurts CRPS.
\begin{table}[htb]
    \caption{Ablation study on \fevbench{} and the subset with future-known covariates.
    The baseline row reports mean SQL skill score $\pm$ std.; other rows report mean deltas
    with original std. over six seeds. Rows below the baseline remove or replace a
    single component of \tirex{-2} (higher is better).}
    \label{tab:ablations}
    \small
    \centering
    \begin{tabular}{@{}lcc@{}}
        \toprule
        Configuration & All \fevbench{} SQL skill / $\Delta$ & Future-known cov. subset SQL skill / $\Delta$ \\
        \midrule
        \tirex{-2} (full) & \textbf{45.5 $\pm$ 0.15} & \textbf{45.19 $\pm$ 0.38} \\
        \midrule
        No group attention & $-$9.67 $\pm$ 2.22 & $-$18.73 $\pm$ 3.44 \\
        No asymm. mask & $-$4.35 $\pm$ 0.24 & $-$8.05 $\pm$ 0.37 \\
        No future-known covariates & $-$1.87 $\pm$ 0.15 & $-$6.59 $\pm$ 0.43 \\
        Forward only & $-$1.37 $\pm$ 0.12 & $-$4.62 $\pm$ 0.37 \\
        sLSTM only & $-$0.26 $\pm$ 0.17 & $-$0.16 $\pm$ 0.28 \\
        No binary-aware scaler & $-$0.13 $\pm$ 0.12 & +0.07 $\pm$ 0.27 \\
        8k$\to$2k context & $-$0.12 $\pm$ 0.13 & $-$0.09 $\pm$ 0.15 \\
        \bottomrule
    \end{tabular}
\end{table}

\begin{table}[h]
  \caption{Leave-one-out ablation of the coupling mechanisms. \fevbench{}:
  skill score against Seasonal Naive over 100 tasks, higher is better, so a
  negative delta means the ablation hurt. GIFT-Eval: raw MASE and CRPS, lower
  is better, so a positive delta means the ablation hurt. Three seeds.}
  \label{tab:coupler-ablation}
  \centering
  \small
  \begin{tabular}{lrrrr}
    \toprule
    & \multicolumn{2}{c}{\fevbench{} skill $\uparrow$} & \multicolumn{2}{c}{GIFT-Eval $\downarrow$} \\
    \cmidrule(lr){2-3}\cmidrule(lr){4-5}
    Coupler removed & SQL & MASE & MASE & CRPS \\
    \midrule
    none (\tirex{-2})     & $45.50 \pm 0.15$ & $33.74 \pm 0.23$ & $0.6970 \pm 0.0006$ & $0.4780 \pm 0.0009$ \\
    \midrule
    Identity/passthrough  & $-2.21 \pm 0.23$ & $-2.62 \pm 0.28$ & $-0.0006 \pm 0.0015$ & $-0.0009 \pm 0.0000$ \\
    Functional            & $-2.09 \pm 0.07$ & $-2.60 \pm 0.04$ & $+0.0066 \pm 0.0040$ & $+0.0016 \pm 0.0006$ \\
    Linear mixing         & $-2.05 \pm 0.08$ & $-2.41 \pm 0.10$ & $+0.0018 \pm 0.0025$ & $+0.0003 \pm 0.0000$ \\
    Cointegration         & $-2.12 \pm 0.05$ & $-2.45 \pm 0.10$ & $+0.0009 \pm 0.0019$ & $+0.0007 \pm 0.0006$ \\
    Linear SCM            & $-2.12 \pm 0.02$ & $-2.48 \pm 0.07$ & $+0.0014 \pm 0.0023$ & $-0.0001 \pm 0.0006$ \\
    Nonlinear SCM         & $-2.97 \pm 0.11$ & $-2.36 \pm 0.20$ & $+0.0000 \pm 0.0007$ & $-0.0002 \pm 0.0006$ \\
    \bottomrule
  \end{tabular}
\end{table}

\newpage

\section{Conclusion}
\label{sec:conclusion}

We presented \tirex{-2}, a recurrent xLSTM-based foundation model for
multivariate time series forecasting with past and future-known covariates. By pairing a
bidirectional time mixer with an asymmetric group-attention variate mixer, \tirex{-2}
exploits future-known covariates while keeping target and past-covariate states strictly
causal, which in turn enables constant-cost streaming inference at arbitrary context lengths.
Together with a synthetic coupling pipeline that supplies the cross-variate diversity missing
from curated multivariate corpora, this yields competitive zero-shot accuracy at a modest
parameter budget: on GIFT-Eval \tirex{-2} improves on the zero-shot Chronos-2-Synth on all
four statistics and on Chronos-2 under a matched corpus, and it wins 65--100\% of \fevbench{}
tasks against every baseline but Chronos-2, with which it splits tasks evenly on SQL
($51.0$, CI $41.0$--$60.0$) while trailing on aggregate skill score. A natural direction for future work is dynamic
covariate selection.
Just as \tirex{-2} bypasses the variate mixer entirely in univariate
mode, an in-context estimate of covariate-target correlation could
prune individual covariates from the group attention whenever
their contribution is negligible, saving FLOPs at inference.

\paragraph{Limitations.} Streaming at constant per-patch cost covers targets,
past covariates, and future-known covariates whose values are available in
advance: since covariate queries never attend to target keys
(\Cref{eq:variate-mask}), the bidirectional covariate representations depend on
covariate inputs alone and are computed once before the stream starts. Only
revisions of already-provided future-known values, e.g.\ an updated weather
forecast, require recomputing their representations and the target states that
read them. This is rarely binding: such covariates (calendars, scheduled
events) are typically fixed, and revisable ones can be shifted into the causal
region as past covariates, which restores full streaming
(\Cref{fig:horizon_scaling_and_shift}). Streaming also operates at patch
granularity (32 steps). Our long-context comparisons run baselines past their
trained context; a sliding window at their native length would narrow the gap,
though the per-patch cost advantage holds.

\section*{Acknowledgments and Disclosure of Funding}
\addcontentsline{toc}{section}{Acknowledgments and Disclosure of Funding}

We thank Sebastian Lehner for extensive discussions throughout the development of the model, and Michael List and Andreas Mayr for their valuable feedback.
We further thank Christian Ganh\"{o}r, Daniil Yefimov, and Martin Loretz for their contributions to the open-source code repository, and Suad Demiri for porting TiRex-2 to Apple devices.
The LIT AI Lab, and the Institute for Machine Learning are supported by the Federal State Upper Austria.
We acknowledge the EuroHPC Joint Undertaking for awarding us access to Leonardo at CINECA (Italy) and MareNostrum 5 at BSC (Spain). 

\bibliographystyle{plainnat}
\bibliography{references}

\clearpage
\appendix
\startcontents[appendix]

\section*{Appendix}
\printcontents[appendix]{l}{1}{\setcounter{tocdepth}{2}}

\newpage

\section{Extended results}
\label{app:experiments}

\begin{figure}[!ht]
  \centering
  \includegraphics[width=0.90\linewidth]{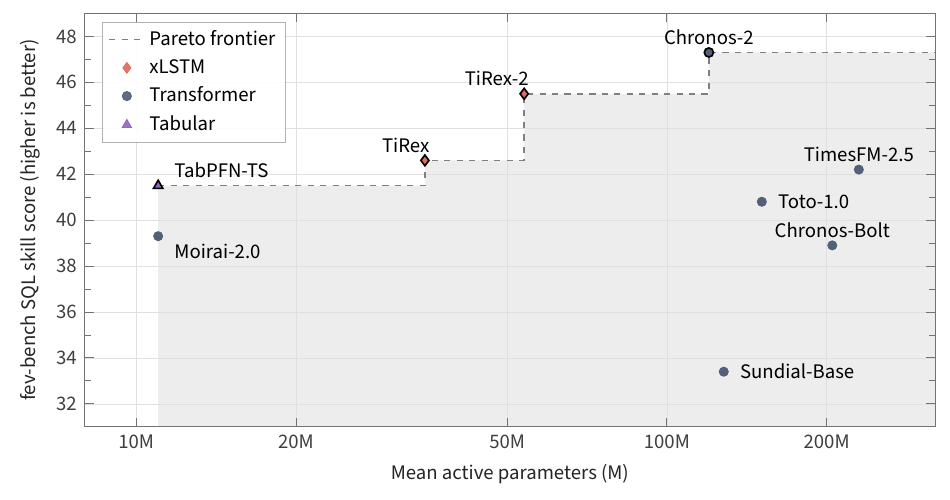}
\caption{%
\fevbench{} SQL skill score against Seasonal Naive versus mean active
parameters across nine zero-shot \timeseries{} foundation models (TSFMs).
Higher is better, so Pareto-optimal models (TabPFN-TS, \tirex{},
\tirex{-2}, Chronos-2) form the upper-left frontier (dashed staircase,
black outlines); the shaded area marks the dominated subset of the
parameter--skill plane. Chronos-2 attains the highest skill score, at
$\sim\!2.2\times$ the mean active parameters of \tirex{-2} and
$\sim\!3\times$ its median runtime ($0.838$ against $0.273$\,s per 100
tasks); \tirex{-2} is the strongest model below 100M active parameters. For
\tirex{-2} the variate mixer is skipped on purely univariate inputs
(Sec.~\ref{sec:model}), so we report the task-weighted expectation
$\dE_t[P_\text{active}] = P_\text{base} + p_\text{cov}\,P_\text{var-mix}
= 38.4 + 0.35\cdot 44.1 \approx 53.84$M over the $65/100$ univariate and
$35/100$ multivariate tasks of \fevbench{}; in a fully multivariate
deployment the count is $82.5$M. All other models are dense, so total and
active parameter counts coincide.
}
\label{fig:tsfm_pareto_skill_vs_params}
\end{figure}

\begin{figure}[!ht]
  \centering
  \includegraphics[width=0.90\linewidth]{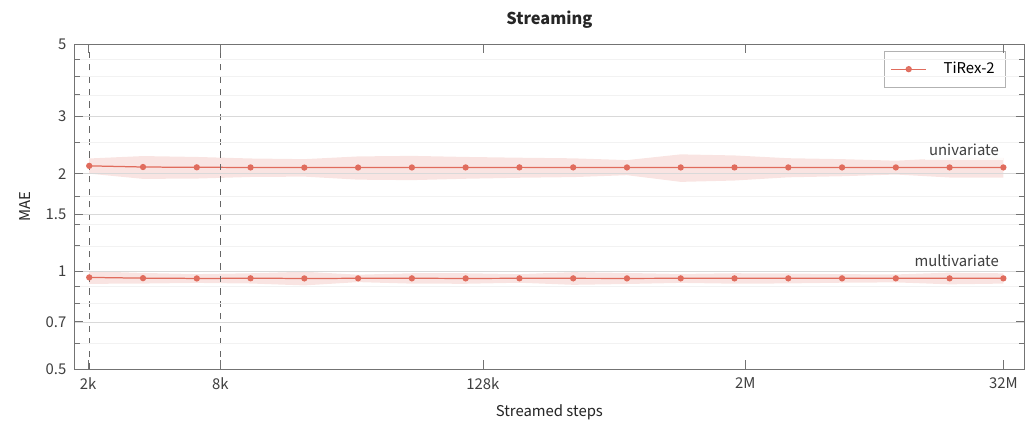}
\caption{%
Streaming evaluation: MAE of \tirex{-2} as a function of cumulative
streamed steps. Univariate and multivariate variants are shown separately.
Shaded regions indicate $\pm 1$ standard deviation of the MAE across
$8$ runs.
Vertical dashed lines mark the pre-training context length ($2$k) and
post-training context length ($8$k).
}\label{fig:streaming_panel_appendix}
\end{figure}

\begin{figure}[p]
  \centering
  \includegraphics[width=0.90\linewidth]{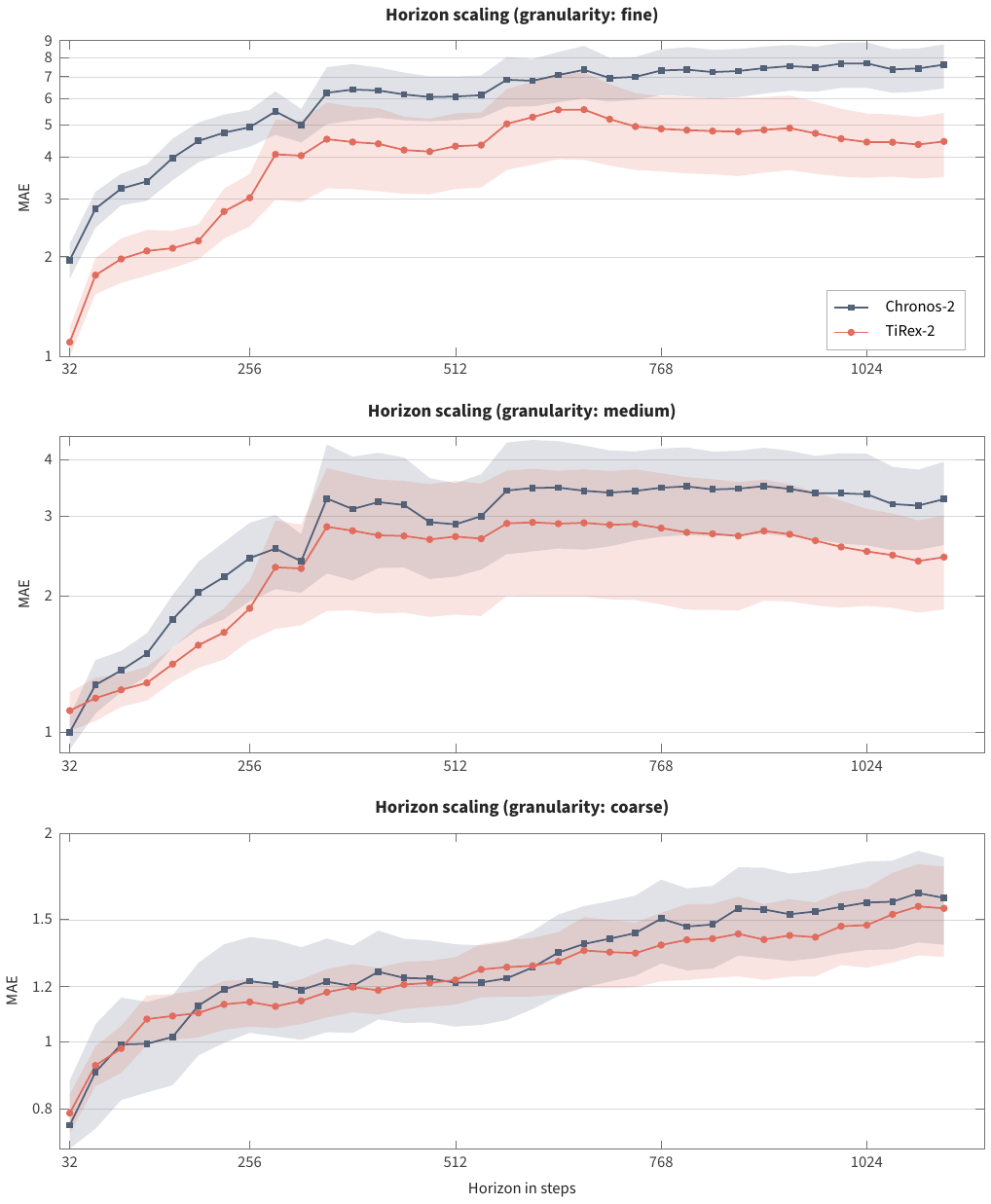}
\caption{%
Long-horizon forecasting on \texttt{dysts}, MAE vs.\ forecast horizon
on a log-$y$ axis. Subplots show fine, medium, and coarse temporal granularity.
Shaded bands indicate $\pm 1$~SEM ($\sigma/\sqrt{N}$, $N{=}135$ tasks).
\tirex{-2} attains lower MAE than Chronos-2 at nearly all horizons. The gap is
largest at fine resolution, where it widens with the horizon; at coarse
resolution the two models are close, with overlapping SEM bands.}
\label{fig:longhorizon_appendix}
\end{figure}

\newpage

\section{Additional experiments and analyses}
\label{app:additional}

\subsection{Reporting under the \fevbench{} protocol}
\label{app:add:fev-protocol}

\Cref{tab:fev-protocol} reports \fevbench{} under the benchmark's own
protocol \citep{shchur_fev-bench_2025}. SQL denotes its scaled quantile loss,
skill scores are relative to Seasonal Naive.

\begin{table}[h!]
  \caption{\fevbench{} (100 tasks) under the benchmark protocol. Win rates are
  pairwise against the other column, with 95\% confidence intervals; skill
  scores are relative to Seasonal Naive. Higher is better throughout.}
  \label{tab:fev-protocol}
  \centering
  \small
  \begin{tabular}{lrr}
    \toprule
    Statistic & \tirex{-2} & Chronos-2 \\
    \midrule
    Pairwise win rate, SQL  & $51.0$ (CI $41.0$--$60.0$) & $49.0$ \\
    Pairwise win rate, MASE & $44.0$ (CI $34.0$--$54.0$) & $\mathbf{56.0}$ \\
    Skill score, SQL        & $45.5$ & $\mathbf{47.3}$ \\
    Skill score, MASE       & $33.7$ & $\mathbf{35.5}$ \\
    \bottomrule
  \end{tabular}
\end{table}

Against the remaining baselines \tirex{-2} wins $65$--$100\%$ of tasks on SQL
and $63$--$99\%$ on MASE, at roughly $2.2\times$ fewer active parameters and
roughly $3\times$ lower median runtime ($0.273$ against $0.838$\,s per 100
tasks).

Broken down by covariate type (\Cref{tab:cov-subset}), \tirex{-2} leads
Chronos-2 on tasks with past-only covariates on both metrics, while Chronos-2
is ahead when future-known covariates are present. On tasks carrying both
types, the three models supporting future-known covariates stay above $37$
SQL and $24$ MASE skill, while all others fall below $31$ and $17$.

\begin{table}[h]
  \caption{\fevbench{} skill scores against Seasonal Naive by covariate type,
  under the benchmark protocol; higher is better, best per column in bold,
  second best underlined. Subsets: all tasks ($n=100$), known covariates only
  ($18$), past covariates only ($12$), both types ($12$), and any dynamic
  covariate ($42$); the middle three partition the last.}
  \label{tab:cov-subset}
  \centering
  \small
  \setlength{\tabcolsep}{4pt}
  \begin{tabular}{lrrrrrcrrrrr}
    \toprule
    & \multicolumn{5}{c}{Skill score, SQL} && \multicolumn{5}{c}{Skill score, MASE} \\
    \cmidrule(lr){2-6}\cmidrule(lr){8-12}
    Model & all & known & past & both & any && all & known & past & both & any \\
    \midrule
    Chronos-2    & $\mathbf{47.3}$ & $\mathbf{52.6}$ & $\underline{42.4}$ & $\mathbf{42.5}$ & $\mathbf{47.0}$
                 && $\mathbf{35.5}$ & $\mathbf{43.2}$ & $36.6$ & $\mathbf{30.3}$ & $\mathbf{37.9}$ \\
    \tirex{-2}   & $\underline{45.5}$ & $\underline{49.7}$ & $\mathbf{43.9}$ & $37.7$ & $\underline{44.8}$
                 && $\underline{33.7}$ & $\underline{39.8}$ & $\mathbf{39.7}$ & $24.6$ & $\underline{35.8}$ \\
    TabPFN-TS    & $41.5$ & $47.1$ & $36.8$ & $\underline{40.6}$ & $42.5$
                 && $29.8$ & $37.7$ & $32.7$ & $\underline{28.1}$ & $33.6$ \\
    TiRex        & $42.6$ & $42.7$ & $40.6$ & $29.9$ & $38.7$
                 && $30.0$ & $31.2$ & $\underline{36.8}$ & $14.1$ & $28.5$ \\
    TimesFM-2.5  & $42.2$ & $41.4$ & $37.4$ & $30.7$ & $37.4$
                 && $30.2$ & $31.0$ & $33.4$ & $16.1$ & $27.8$ \\
    Toto-1.0     & $40.8$ & $37.7$ & $39.0$ & $26.9$ & $35.2$
                 && $28.2$ & $26.2$ & $34.0$ & $12.8$ & $25.1$ \\
    FlowState    & $39.9$ & $40.3$ & $36.8$ & $30.5$ & $36.7$
                 && $27.8$ & $30.5$ & $32.4$ & $15.9$ & $27.2$ \\
    Chronos-Bolt & $38.9$ & $40.9$ & $34.9$ & $28.6$ & $35.9$
                 && $26.5$ & $30.3$ & $29.5$ & $13.5$ & $25.6$ \\
    \bottomrule
  \end{tabular}
\end{table}

\subsection{GIFT-Eval under a matched training corpus}
\label{app:add:gift-matched}

Two \tirex{-2} checkpoints are evaluated: one strictly zero-shot, with all
overlapping datasets removed (\Cref{app:pretrain-corpus}), and one trained on a
corpus matched to Chronos-2. Neither configuration touches the test split.

\begin{table}[h]
  \caption{GIFT-Eval. Lower is better throughout; ranks are averaged over
  tasks. Chronos-2 trains on the GIFT-Eval pre-train splits, so its zero-shot
  counterpart is Chronos-2-Synth; comparisons are meaningful only within a
  block. No configuration touches the test split.}
  \label{tab:gift-matched}
  \centering
  \small
  \begin{tabular}{lrrrr}
    \toprule
    Model & MASE & MASE rank & CRPS & CRPS rank \\
    \midrule
    \multicolumn{5}{l}{\emph{Zero-shot: GIFT-Eval pre-train splits unseen}} \\
    \tirex{-2} (zero-shot) & $\mathbf{0.697}$ & $\mathbf{43.18}$ & $\mathbf{0.478}$ & $\mathbf{38.28}$ \\
    Chronos-2-Synth        & $0.720$ & $60.47$ & $0.496$ & $59.28$ \\
    \midrule
    \multicolumn{5}{l}{\emph{Trained on the GIFT-Eval pre-train splits}} \\
    \tirex{-2} (matched corpus) & $\mathbf{0.678}$ & $\mathbf{36.20}$ & $\mathbf{0.467}$ & $\mathbf{34.68}$ \\
    Chronos-2                   & $0.698$ & $38.96$ & $0.485$ & $39.50$ \\
    \bottomrule
  \end{tabular}
\end{table}

\subsection{Streaming: per-patch latency and memory}
\label{app:add:latency}
At batch size~1, Chronos-2 costs $17.8$\,ms up to 2k, $107$\,ms at
$2.6\times10^{4}$ steps, $1.67$\,s at $2.1\times10^{5}$ and $28.4$\,s at
$1.05$M; \tirex{-2} recompute costs $206$\,ms at $9.7\times10^{4}$,
$956$\,ms at $4.7\times10^{5}$ and $2.07$\,s at $1.05$M. Streaming costs
$15.48$--$15.60$\,ms throughout, i.e.\ at $1.05$M steps $133\times$ less than recomputation and $1830\times$
less than Chronos-2, which has no incremental path.
Streamed memory is $0.318$\,GB at batch size~1 and $0.42$\,GB at batch
size~64 at all contexts. At batch size~64 recompute reaches $28.8$\,GB
(Chronos-2, $2.0\times10^{5}$ steps) and $36.9$\,GB (\tirex{-2},
$4.7\times10^{5}$ steps).

\subsection{Streaming: covariate contribution}
\label{app:add:streaming-eff}

Chronos-2 is evaluated under the same streaming regime with its 8k truncation
removed, so that it sees the full stream.

\begin{table}[h]
  \caption{Streaming to extrapolated context lengths on the AR(1) stream of
  \Cref{app:add:setup}, mean MAE $\pm$ 1 standard deviation across eight runs. UV is univariate, MV adds one informative
  covariate; lower is better. A value is bold only where the two models'
  $\pm 1$ SD intervals do not overlap.}
  \label{tab:chronos-streaming}
  \centering
  \small
  \begin{tabular}{lrrrr}
    \toprule
    & \multicolumn{2}{c}{Chronos-2} & \multicolumn{2}{c}{\tirex{-2}} \\
    \cmidrule(lr){2-3}\cmidrule(lr){4-5}
    Context & UV & MV & UV & MV \\
    \midrule
    2k   & $2.110 \pm 0.981$ & $0.766 \pm 0.192$ & $2.114 \pm 0.322$ & $0.959 \pm 0.121$ \\
    8k   & $2.857 \pm 1.336$ & $1.214 \pm 0.490$ & $2.092 \pm 0.452$ & $0.952 \pm 0.084$ \\
    32k  & $2.614 \pm 1.570$ & $2.588 \pm 1.446$ & $2.088 \pm 0.485$ & $\mathbf{0.952 \pm 0.071}$ \\
    128k & $2.453 \pm 0.792$ & $2.456 \pm 0.806$ & $2.090 \pm 0.456$ & $\mathbf{0.952 \pm 0.101}$ \\
    \bottomrule
  \end{tabular}
\end{table}

The UV\,--\,MV gap of \tirex{-2} runs from $1.155$ at 2k to $1.139$ at 128k;
that of Chronos-2 from $1.344$ at 2k to $0.026$ at 32k and $-0.003$ at 128k.

\subsection{Architectural ablations on GIFT-Eval}
\label{app:add:gift-ablation}

\begin{table}[h]
    \caption{Architectural ablations on GIFT-Eval. First row absolute, others
      deltas against it; lower is better, so a positive delta means the ablation
      hurt. Three seeds (\Cref{tab:ablations}: six). GIFT-Eval contains far
      fewer tasks with informative cross-variate structure than \fevbench{}, so the
      group-attention delta is diluted and not comparable to \Cref{tab:ablations}.}
  \label{tab:gift-ablation}
  \centering
  \small
  \begin{tabular}{lrr}
    \toprule
    Variant & MASE & CRPS \\
    \midrule
    \tirex{-2} (full)          & $\mathbf{0.697} \pm 0.0006$ & $\mathbf{0.478} \pm 0.0009$ \\
    \midrule
    w/o group attention        & $+0.0018 \pm 0.0027$ & $+0.0032 \pm 0.0028$ \\
    w/o binary-aware scaler    & $-0.0013 \pm 0.0011$ & $+0.0000 \pm 0.0012$ \\
    sLSTM only                 & $+0.0019 \pm 0.0020$ & $+0.0022 \pm 0.0007$ \\
    Context length 2k          & $+0.0156 \pm 0.0011$ & $+0.0111 \pm 0.0012$ \\
    \bottomrule
  \end{tabular}
\end{table}

\Cref{tab:gift-ablation} shows architecture ablations evaluated on GIFT-Eval. 
Relative to the pure sLSTM backbone, the mixed backbone gives a median
inference speed-up of $5$ to $7\%$, up to roughly $20\%$ on the smallest
datasets, and trains $25\%$ faster. Removing the binary-aware bypass costs
$0.4 \pm 0.1$ SQL skill points on the 43 short-context \fevbench{} tasks,
against $0.1 \pm 0.1$ overall.

\subsection{Scaling the number of covariates}
\label{app:add:cov-scaling}

In \Cref{tab:cov-scaling-128,tab:cov-scaling-256} we contrast TiRex-2's and
Chronos-2's ability of extracting progressively more information from future
known covariates.
We use $M$ independent AR(1) series whose scaled sum, perturbed by Gaussian noise,
forms the target, with $k$ of them revealed as known past and future
covariates (\Cref{app:add:setup}).

\begin{table}[h]
  \caption{Covariate scaling at $M = 128$, construction in
  \Cref{app:add:setup}. MSE of the median forecast over 128 repetitions
  (SEM $\approx 0.02$) and per-call latency in milliseconds; lower is better
  for both.}
  \label{tab:cov-scaling-128}
  \centering
  \small
  \begin{tabular}{lrrrr}
    \toprule
    $k$ & \tirex{-2} MSE & Chronos-2 MSE & \tirex{-2} ms & Chronos-2 ms \\
    \midrule
    $0$   & $1.080$ & $1.080$ & $\mathbf{0.78}$  & $1.15$ \\
    $16$  & $1.056$ & $\mathbf{1.031}$ & $\mathbf{9.71}$  & $17.36$ \\
    $64$  & $\mathbf{0.913}$ & $0.998$ & $\mathbf{44.50}$ & $63.35$ \\
    $128$ & $\mathbf{0.819}$ & $0.996$ & $\mathbf{66.20}$ & $135.97$ \\
    \bottomrule
  \end{tabular}
\end{table}

\begin{table}[h]
  \caption{Covariate scaling at $M = 256$, MSE of the median forecast
  (mean $\pm$ 1 SEM). Larger $M$ lowers each covariate's individual
  contribution, so the two sweeps are not directly comparable at equal $k$.}
  \label{tab:cov-scaling-256}
  \centering
  \small
  \begin{tabular}{lrr}
    \toprule
    $k$ & \tirex{-2} MSE & Chronos-2 MSE \\
    \midrule
    $0$   & $1.093 \pm 0.021$ & $1.095 \pm 0.021$ \\
    $16$  & $1.104 \pm 0.023$ & $1.114 \pm 0.020$ \\
    $32$  & $1.079 \pm 0.022$ & $1.073 \pm 0.020$ \\
    $96$  & $\mathbf{0.991 \pm 0.020}$ & $1.059 \pm 0.019$ \\
    $256$ & $\mathbf{0.932 \pm 0.018}$ & $1.057 \pm 0.019$ \\
    \bottomrule
  \end{tabular}
\end{table}

At $M = 256$ the $96 \to 256$ step still yields a gain for \tirex{-2};
Chronos-2 is flat from $k \approx 32$ at $M = 128$ and $k \approx 96$ at
$M = 256$. In units of covariates' worth of information ($\Delta \cdot M$),
\tirex{-2} extracts 33 to 41 against Chronos-2's roughly 10. At $M = 1024$ the
per-covariate correlation with the target is $0.030$ ($0.085 \to 0.060 \to
0.030$ across the three sweeps) and neither model improves over $k = 0$; a
$k = 1024$ forward pass costs \tirex{-2} $0.52$\,s and $4.0$\,GB against
Chronos-2's $2.28$\,s and $21.9$\,GB. 

\subsection{Experimental setup for the streaming, ablation and covariate-scaling results}
\label{app:add:setup}

This section collects the experimental setups behind the results reported in
\Crefrange{app:add:latency}{app:add:cov-scaling}, one paragraph per
subsection and in the same order. Every trained arm in this appendix uses three
seeds and we report mean $\pm$ std across them; \Cref{tab:ablations} in the main
text uses up to six for some arms, and each caption states its own count.

\paragraph{Latency and memory (\Cref{app:add:latency}).}
Single NVIDIA A40 with 48\,GB, one process on one GPU. Both models run in
float32 with \texttt{torch.set\_float32\_matmul\_precision("highest")}. One
series, context from 192 steps to $1.05\times10^{6}$, at batch size 64 and 1;
a random-walk series with a fixed seed, so that context length is exactly
controllable. We perform 200 timed calls per configuration after 32 untimed
warmups, each call bracketed by \texttt{torch.cuda.synchronize()} and timed
with \texttt{time.perf\_counter()}, and report the mean end-to-end predict
latency; run-to-run standard deviation is at most $0.45$\,ms throughout. A
configuration is dropped once it exhausts the device. Chronos-2 uses patch
size 16; \tirex{-2} uses patch size 32, so one streamed step
ingests one new patch. Latency is reported per call, not per forecast step.

\paragraph{Streaming covariate contribution (\Cref{app:add:streaming-eff};
\Cref{tab:chronos-streaming} and \Cref{fig:horizon_scaling_and_shift}).}
A first-order autoregressive target driven by a noisy, lagged covariate: the
target responds to the covariate with a delay, so the covariate's relevant
information always lies in the target's relative past. The covariate is
registered as future-known. Both are ingested patch by patch with a forecast
emitted after every update; eight runs per configuration.

\paragraph{Architectural ablations (\Cref{app:add:gift-ablation}; \Cref{tab:coupler-ablation,tab:gift-ablation}).}
Each arm is trained for 700k steps on three seeds, with architecture and
evaluation configuration held fixed across arms. \fevbench{} results follow
the benchmark protocol \citep{shchur_fev-bench_2025} with skill scores against
Seasonal Naive over 100 tasks; GIFT-Eval results are raw MASE and CRPS under
that benchmark's protocol \citep{aksu_gift-eval_2024}.

\paragraph{Covariate scaling (\Cref{app:add:cov-scaling}; \Cref{tab:cov-scaling-128,tab:cov-scaling-256}).}
We generate $M$ independent AR(1) series $x_{i,t}$ with $\rho = 0.9$, each
standardized to unit marginal variance via innovation variance $1 - \rho^2$.
Their sum scaled by $1/\sqrt{M}$ is a unit-variance signal, which we perturb
with Gaussian noise of $\sigma = 0.3$ to form the target $y$; $\sigma^2 = 0.09$
is an irreducible error floor. Revealing $k$ of the $M$ series leaves a
Bayes-optimal error of $\sigma^2 + (M-k)/M$. We supply $k$ series as known past
and future covariates, nested across $k$, with context 2048 and horizon 320,
and score the median forecast by MSE over 128 repetitions, reporting mean
$\pm$ 1 SEM. Both models see identical realizations, so the comparison is
paired.

\newpage
\section{Asymmetric group attention: leakage derivation and details}
\label{app:asymmetric-mask}

This appendix expands on the variate-mixer paragraph of \Cref{sec:model} and gives (i) the full specification of the asymmetric group attention, (ii) a proof that the
asymmetric mask together with the forward-only \xlstm{} on targets and past covariates is \emph{sufficient} for
target-causality (\Cref{prop:target-causality}), and (iii) a counterexample showing that the asymmetric mask is
\emph{necessary}: dropping the $\cY\cup\cZ\!\to\!\cX$ block opens a concrete two-block leakage path
from future targets into earlier target representations.

\subsection{Notation and group structure}
As in the main paper, a mixer is a map $\BH\mapsto\BH'$ between token tensors in $\reals^{F\times L\times D}$, and one
\tirex{-2} block applies a time mixer followed by a variate mixer. We write $\Bh_{f,l}$ for the token of variate $f$ at
patch $1\leq l\leq L$, and $\cX,\cY,\cZ\subseteq\{1,\dots,F\}$ for the index sets of the target,
past-covariate and future-covariate variates. Depth enters only through block outputs: as in \Cref{lem:one-block},
$\BH^{n}$ is the output of the $n$-th block and $\BH^{0}$ the output of the input layer; representations inside a
block carry no index.
The variates are partitioned into groups. Inside the variate mixer we transpose to
shape $L\times F\times D$ and apply attention along the $F$ axis independently per patch position $l$.
The additive mask $M\in\{0,-\infty\}^{F\times F}$ is shared across all patch positions and block-diagonal in
this group structure: every entry linking variates of different groups is set to $-\infty$, while within a group
\begin{equation}
  m_{ij} \;=\;
  \begin{cases}
    0      & \text{if } \neg\!\bigl(i\!\in\!\cY\cup\cZ\,\wedge\,j\!\in\!\cX\bigr),\\
    -\infty & \text{otherwise,}
  \end{cases}
  \label{eq:app:mask}
\end{equation}
where $i,j\in\{1,\dots,F\}$ index (query,\,key). Equivalently, within a group the mask sends each query to its set of admissible keys via
\begin{equation*}
  i\;\longmapsto\;
  \begin{cases}
    \cX\cup\cY\cup\cZ & \text{if } i\in\cX,\\
    \cY\cup\cZ                & \text{if } i\in\cY\cup\cZ,
  \end{cases}
\end{equation*}
so target queries attend to target and covariate keys, whereas covariate queries attend to covariate keys only.

\subsection{Proof of \cref{prop:target-causality} (sufficiency)}
\label{app:proof-of-causality}

This subsection establishes that the asymmetric mask~\eqref{eq:variate-mask} together with the forward-only
\xlstm{} on targets and past covariates is sufficient to keep target tokens free of future-target dependencies
at every depth; necessity of the mask is treated separately in \Cref{app:leakage}.

\paragraph{Dependency sets.}
For a token $\Bh_{f,l}$ of any representation in the stack, let
$\mathrm{Dep}(\Bh_{f,l})\subseteq\{1,\dots,F\}\times\{1,\dots,L\}$
be the set of pairs $(f',l')$ such that $\Bh_{f,l}$ is a non-trivial function of the input token $\Bh^{0}_{f',l'}$.
Define the temporal receptive field $I_f(l)=\{1,\dots,L\}$ if $f\in\cZ$ and $I_f(l)=\{1,\dots,l\}$ otherwise,
and the variate receptive set $S_{f}=\cX\cup\cY\cup\cZ$ for $f\in\cX$ and $S_{f}=\cY\cup\cZ$ for $f\in\cY\cup\cZ$.
$I_f$ encodes the directionality of the time mixer (forward-only on targets and past covariates, bidirectional on
future covariates), and $S_f$ encodes the asymmetric mask~\eqref{eq:variate-mask} (covariate queries cannot read
target keys). For a mixer $\BH\mapsto\BH'$ this gives
\begin{align*}
  \mathrm{Dep}\bigl(\Bh'_{f,l}\bigr)
    &\subseteq \!\!\bigcup_{l'\in I_f(l)}\!\! \mathrm{Dep}\bigl(\Bh_{f,l'}\bigr)
    && \text{(time mixer),}\\
  \mathrm{Dep}\bigl(\Bh'_{f,l}\bigr)
    &\subseteq \bigcup_{f'\in S_f} \mathrm{Dep}\bigl(\Bh_{f',l}\bigr)
    && \text{(variate mixer),}
\end{align*}
where the variate-mixer union is at fixed patch $l$ since the mixer acts independently per patch. Normalization,
the token-wise MLPs and the residual connections act on each token separately and add nothing: the residual term
contributes $\mathrm{Dep}(\Bh_{f,l})$, which both unions already contain since $l\in I_f(l)$ and $f\in S_f$.

\paragraph{Invariant.}
Call a representation \emph{admissible} if its tokens satisfy
\begin{align*}
  \mathbf{(C)}\;\; & f\in\cX,\;(f',l')\in \mathrm{Dep}\bigl(\Bh_{f,l}\bigr)\;\text{ and }\;f'\in\cX\;\Longrightarrow\;l'\leq l,\\
  \mathbf{(I)}\;\; & f\in\cY\cup\cZ\;\Longrightarrow\;\mathrm{Dep}\bigl(\Bh_{f,l}\bigr)\cap(\cX\times\{1,\dots,L\})=\emptyset,
\end{align*}
i.e.\ target tokens depend on no later target input, and covariate tokens depend on no target input at all.
We show that the input is admissible and that every mixer maps an admissible representation $\BH$ to an admissible $\BH'$.

\emph{Input layer}: $\mathrm{Dep}(\Bh^{0}_{f,l})=\{(f,l)\}$ satisfies both.
\emph{Time mixer}: for $f\in\cX$, $I_{f}(l)\subseteq\{1,\dots,l\}$,
so every target-index pair inherited from $\BH$ has $l'\leq l$, preserving $\mathbf{(C)}$. For $f\in\cY\cup\cZ$,
the union runs over tokens of the same covariate, and $\mathbf{(I)}$ carries over unchanged.
\emph{Variate mixer}: for $f\in\cX$, the union runs over $f'\in S_{f}=\cX\cup\cY\cup\cZ$ at fixed patch $l$;
contributions from keys $f'\in\cX$ satisfy $\mathbf{(C)}$ in $\BH$ (so any target-index pair has $l'\leq l$),
while contributions from $f'\in\cY\cup\cZ$ contain no target-index pair by $\mathbf{(I)}$, preserving $\mathbf{(C)}$.
For $f\in\cY\cup\cZ$, $S_f=\cY\cup\cZ$ excludes $\cX$, and $\mathbf{(I)}$ holds in $\BH$ for every $f'\in S_f$, so $\mathbf{(I)}$ is preserved.

Applying this to the $2N$ mixers of the stack in order, every representation in the stack is admissible, in
particular every block output $\BH^{n}$, $n\leq N$. Property $\mathbf{(C)}$ for $\BH^{n}$ is
\Cref{prop:target-causality}.\hfill$\square$

Intuitively, $\mathbf{(I)}$ is what makes the bidirectional time mixer on future covariates safe: covariate tokens never accumulate target information,
so the reverse pass cannot transport target content from a later patch back to an earlier one.

\subsection{Necessity of the asymmetric mask: two-block leakage counterexample}
\label{app:leakage}
The sufficiency proof in \Cref{app:proof-of-causality} leaves open whether the $\cY\cup\cZ\!\to\!\cX$
block in~\Cref{eq:app:mask} is actually needed, or whether a symmetric variate mixer would already preserve
target-causality. We resolve this by exhibiting an explicit leakage path: without the
$\cY\cup\cZ\!\to\!\cX$ block, a target representation at patch $l$ depends on target inputs at
positions $l'>l$, even within a single forward pass and without any autoregressive sampling. The leakage path uses one symmetric variate mixer, the bidirectional time
mixer of the next block, and a second variate mixer; three components that are individually well-defined but, in
combination, route information backwards along the time axis.

Assume the series contains at least one future-covariate variate (index $z\in\cZ$) and one
target variate (index $x\in\cX$) in the same group, and follow three consecutive mixers, the primes counting the
mixers applied so far:
\begin{equation*}
  \BH\;\xrightarrow{\text{variate mixer}}\;\BH'\;\xrightarrow{\text{time mixer}}\;\BH''\;\xrightarrow{\text{variate mixer}}\;\BH'''.
\end{equation*}

\paragraph{Step 1 (symmetric variate mixer).}
With a symmetric mask, the future-covariate token at position $l$ aggregates over all variates in the
group, including the target variate $x$:
\begin{equation*}
  \Bh'_{z,l} \;=\;
  \Bh_{z,l} + \sum_{v\in\cX\cup\cY\cup\cZ} \alpha_{z,v,l}\,\Bh_{v,l}
   \quad\text{with } \alpha_{z,x,l}\neq 0 \text{ in general.}
\end{equation*}
Hence $\Bh'_{z,l}$ carries information about $\Bh_{x,l}$.

\paragraph{Step 2 (bidirectional time mixer, next block).}
The future-covariate stream is processed by both a forward and a reverse xLSTM. The reverse pass
writes into future-covariate tokens at positions $l'<l$ a representation that is a function of all
$\Bh'_{z,l''}$ with $l''\geq l'$, including $l''=l$:
\begin{equation*}
  \Bh''_{z,l'} \;=\;
  \phi\!\PAR{\Bh'_{z,l'},\,\Bh'_{z,l'+1},\,\ldots,\,\Bh'_{z,L}},
\end{equation*}
where $\phi$ collects the forward output, the reverse output, and their linear fusion. By Step 1, $\phi$ is a
function of $\Bh_{x,l}$, the target token at the \emph{later} position $l$.

\paragraph{Step 3 (variate mixer, next block).}
With a symmetric variate mixer, the target token at position $l'<l$ attends to all variates of
its group at the same patch index, including the future covariate $z$:
\begin{equation*}
  \Bh'''_{x,l'} \;=\; \Bh''_{x,l'} + \alpha'_{x,z,l'}\,\Bh''_{z,l'} + \sum_{v\neq z}\alpha'_{x,v,l'}\,\Bh''_{v,l'},
\end{equation*}
and by Step 2 the middle term is a function of $\Bh_{x,l}$ with $l>l'$.
Composing the three steps, the target representation at the earlier patch $l'$ is a non-trivial function of the
target input at the later patch $l$. During training this leak corrupts the supervision signal: the loss at position $l'$ can be reduced by copying from
position $l>l'$, and at inference (where future targets are absent) the resulting representation distribution is
shifted away from training. Single-pass streaming forecasts are then no longer well-defined either.

\paragraph{How the asymmetric mask breaks the chain.}
The mask in Eq.~\eqref{eq:app:mask} forbids exactly the
$\cY\cup\cZ\!\to\!\cX$ direction by setting $\alpha_{z,x,l}=0$ for
all $l$, thus $\Bh'_{z,l}$ is independent of any target token, and steps 2--3
cannot inject target information at later positions back into earlier target tokens. 

\subsection{Comparison to prior cross-variate asymmetric attention designs}
\label{app:prior-cross-variate}
Several recent models also impose an asymmetric target$\leftrightarrow$covariate information flow, but
in different regimes. TimeXer~\citep{wang_timexer_2024} compresses
each covariate series into a single learnable token and lets target queries
cross-attend to those tokens. \tirex{-2}, instead, attends at every patch position,
preserving the temporal alignment between target and covariate features. TimeXer
is moreover a task-specific model and is not evaluated as a zero-shot foundation
model. Timer-XL~\citep{liu_timer-xl_2024} flattens variates and time
into a single token sequence and combines causal intra-variate attention with
covariate-asymmetric attention inside one block. Because it is fully causal
along time, it cannot exploit future covariates. Our design avoids this
limitation by treating future covariates bidirectionally in the time mixer.
CITRAS~\citep{yamaguchi_citras_2025} likewise factorises time and
variate attention and forms variate attention with the target as query and
covariates as keys/values. For future covariates it pairs key at patch $l$ with
value at patch $l{+}1$ to look one step ahead. CITRAS is again task-specific
and its one-step shift is a heuristic substitute
for the proper bidirectional treatment of future covariates that we adopt.

\subsection{Comparison to multivariate and covariate-aware TSFMs}
\label{app:tsfm-comparison}

Table~\ref{tab:tsfm-comparison} positions \tirex{-2} against existing
multivariate \timeseries{} foundation models and covariate-aware single-target
models along three axes: (i)~whether the model can ingest past- and (ii)~future  covariates, and
(iii)~whether its architecture preserves \emph{target causality}, i.e.,\ whether
target representations at patch $l$ are guaranteed to be independent of target
inputs at patches $l'>l$. To our knowledge \tirex{-2} is the first TSFM to provide support for future covariates and stay causal on the target variate(s). This is a prerequisite for the streaming inference of
Sec.~\ref{subsec:streaming} since otherwise target states at earlier patches
would change as new \timestep{}s arrive.

\begin{table}[h]
  \centering
  \small
  \caption{Past and future-covariate support and target-causality preservation across
  multivariate and covariate-aware \timeseries{} foundation models.
  $\checkmark$~=~supported/preserved, $\times$~=~not supported/not preserved.}
  \label{tab:tsfm-comparison}
  \newcommand{\cmark}{$\checkmark$}%
  \newcommand{\xmark}{$\times$}%
  \begin{tabular}{lccc}
    \toprule
    Model & Past covariates & Future covariates & Target causality \\
    \midrule
    \tirex{-2} (ours)                             & \cmark & \cmark & \cmark \\
    \midrule
    \multicolumn{4}{l}{\emph{Multivariate TSFMs}} \\
    \moirai{}~\citep{woo_unified_2024}            & \cmark & \cmark & \xmark \\
    \moiraimoe{}~\citep{liu_moirai-moe_2025}      & \cmark & \xmark & \xmark \\
    Toto~\citep{cohen_this_2025}                  & \cmark & \xmark & \cmark \\
    Chronos-2~\citep{ansari_chronos-2_2025} & \cmark & \cmark & \xmark \\
    TabPFN-TS~\citep{hoo_tables_2026} & \xmark & \cmark & \xmark \\
    GTT~\citep{feng_general_2024}                 & \cmark & \xmark & \xmark \\
    \morpheus{}~\citep{patil_morpheus_2025}       & \cmark & \cmark & \xmark \\
    \midrule
    \multicolumn{4}{l}{\emph{Covariate-aware single-target TSFMs}} \\
    COSMIC~\citep{auer_zero-shot_2025}            & \cmark & \cmark & \xmark \\
    TimesFM-ICF~\citep{faw_-context_2025} & \cmark & \xmark & \cmark \\
    
    \bottomrule
  \end{tabular}
\end{table}

\newpage
\section{Binary-aware tail-compressing scaler}
\label{app:scaler}
This appendix expands on the input-layer paragraph in Sec.~\ref{sec:model} and gives (i) the motivation for the
binary-aware bypass, (ii) the full forward and inverse transforms with numerical clipping, and (iii) the per-variate
statistics used by the scaler.

Given a single variate $\Bx \in \reals^T$ with index set of observed entries
$\cV \subseteq \{1, \dots, T\}$, we compute per-variate statistics over the full context window:
\begin{equation}
  \hat{\mu} = \frac{1}{|\cV|} \sum_{t \in \cV} x_t, \qquad
  \hat{\sigma} = \max\!\Bigg(\sqrt{\frac{1}{|\cV|} \sum_{t \in \cV}
    (x_t - \hat{\mu})^2},\; \epsilon\Bigg),
\end{equation}
where $\epsilon > 0$ is a small constant that prevents division by zero for near-constant or sparse variates.

\paragraph{Why binary signals need a bypass.}
Naive standardization is problematic for sparse binary signals. Consider the
rare-positive case $\bar{p} \ll 1$ (the rare-negative case $\bar{p} \to 1$ is
symmetric by exchanging the roles of the two levels): for $X \in \{0,1\}$ with
$\dP(X=1) = \bar{p}$, the empirical statistics become $\hat{\mu} \approx
\bar{p}$ and $\hat{\sigma} \approx \sqrt{\bar{p}(1-\bar{p})} \approx
\sqrt{\bar{p}}$. The two standardized levels are then
\begin{equation}
  z_0 \;=\; \frac{0 - \hat{\mu}}{\hat{\sigma}} \;\approx\; \frac{-\bar{p}}{\sqrt{\bar{p}}} \;=\; -\sqrt{\bar{p}},
  \qquad
  z_1 \;=\; \frac{1 - \hat{\mu}}{\hat{\sigma}} \;\approx\; \frac{1 - \bar{p}}{\sqrt{\bar{p}}} \;=\; \frac{1}{\sqrt{\bar{p}}} - \sqrt{\bar{p}},
\end{equation}
yielding a gap
\begin{equation}
  z_1 - z_0 \;\approx\; \frac{1}{\sqrt{\bar{p}}} - \sqrt{\bar{p}} - \PAR{-\sqrt{\bar{p}}}
  \;=\; \frac{1}{\sqrt{\bar{p}}}
\end{equation}
that grows without bound as $\bar{p} \to 0$. The representation of both levels
therefore depends on the context-window sparsity rather than the binary
semantics of the signal. To avoid this, the affine transformation is bypassed
for detected binary variates:
\begin{equation}
  \mu = (1 - b)\,\hat{\mu}, \qquad
  \sigma = (1 - b)\,\hat{\sigma} + b, \qquad
  b = \BOn\!\Big[\,\forall\, t \in \cV:\; x_t \in \{0, 1\}\Big].
\end{equation}
This gated parameterization replaces $(\hat{\mu}, \hat{\sigma})$ with $(0, 1)$
when $b=1$, so the affine transform reduces to the identity without branching.
The construction preserves the canonical $\{0,1\}$ encoding and yields a stable,
sparsity-invariant input regardless of the class balance observed in the context.

\paragraph{Forward and inverse transforms.}
The forward transformation standardizes and applies an $\operatorname{arcsinh}$
tail compression to non-binary variates while leaving binary ones unchanged;
the inverse undoes the compression and rescales back to the original domain:
\begin{equation}
  \tilde{x}_t = (1 - b)\,\operatorname{arcsinh}\!\!\PAR{\tfrac{x_t - \mu}{\sigma}}
    + b\, x_t,
  \qquad
  \hat{x}_t = (1 - b)\,\Big(\sigma\, \operatorname{sinh}\!\big(
    \operatorname{clip}_c(\tilde{x}_t)\big) + \mu\Big)
    + b\, \tilde{x}_t,
\end{equation}
where $\operatorname{clip}_c(z) = \max(-c, \min(c, z))$ bounds the input to
$\sinh$ to prevent its exponential growth from exceeding the representable
range of the output datatype and to suppress implausibly large outlier
predictions. We use a fixed bound $c = 20$ in training and at inference. Since
the outputs of the model in the normalized domain typically stay well below
this bound, the clip is only very rarely active in practice and acts mainly as
a safeguard against overflow, as $\sinh(20) \approx 2.4 \times 10^8$ is far
below the range of the output datatype.
A drop-in alternative is to compute $c$ per sample from the per-variate statistics $(\mu, \sigma)$ as
\begin{equation}
  c \;=\; \alpha \cdot \operatorname{arcsinh}\!\PAR{\frac{x_{\max}^{(\mathrm{dtype})} - \mu}{\sigma}},
\end{equation}
with safety factor $\alpha \in (0,1]$, which guarantees $\sigma \sinh(c) + \mu
\leq x_{\max}^{(\mathrm{dtype})}$ and, for $\alpha < 1$, additionally suppresses
implausibly large outlier predictions. In our experiments this alternative
yields virtually identical results, since model outputs typically stay well
below either bound and the clip is only very rarely active.

\newpage
\section{Training setup}
\label{app:training-setup}

We pretrain \tirex{-2} for $700{,}000$ optimizer steps on $2$ NVIDIA H100 GPUs in
bf16-mixed precision. Training takes approximately $50$ hours wall-clock time at
an effective batch size of $64$ (per-GPU batch size of $32$, no gradient
accumulation). Optimization uses AdamW \citep{loshchilov_decoupled_2019} with a
weight decay of $0.01$ and gradients are value-clipped at $1.0$ to stabilize the
update magnitude in the presence of the heavy-tailed sample-loss distribution
typical of large-scale time-series corpora.

\paragraph{Learning-rate schedule.}
The learning rate follows the OneCycle schedule \citep{smith_super-convergence_2019} as
implemented by \texttt{torch.optim.lr\_scheduler.OneCycleLR} with cosine
annealing in both phases \newline (\texttt{anneal\_strategy='cos'}). The schedule is
parameterized by a peak learning rate $\eta_{\max} = 1.2 \cdot 10^{-3}$, an
initial divisor $d_0 = 50$ and a final divisor $d_f = 10^{4}$, which fix the
boundary values
\begin{equation}
  \eta_0 \;=\; \frac{\eta_{\max}}{d_0} \;=\; 2.4 \cdot 10^{-5},
  \qquad
  \eta_f \;=\; \frac{\eta_0}{d_f} \;=\; \frac{\eta_{\max}}{d_0 \, d_f}
       \;=\; 2.4 \cdot 10^{-9}.
\end{equation}
With total training steps $T = 700{,}000$ and warmup fraction
$\rho = 0.05$ (\texttt{pct\_start}), the warmup phase ends at step
$T_w = \rho T = 35{,}000$. Letting $t$ denote the current optimizer step and
defining the normalized phase progress
\begin{equation}
  s_w(t) \;=\; \frac{t}{T_w},
  \qquad
  s_a(t) \;=\; \frac{t - T_w}{T - T_w},
\end{equation}
the schedule is given by
\begin{equation}
  \eta(t) \;=\;
  \begin{cases}
    \eta_{\max} + \tfrac{1}{2}\big(\eta_0 - \eta_{\max}\big)\big(1 + \cos(\pi\, s_w(t))\big),
      & 0 \leq t \leq T_w \quad \text{(cosine warmup)}, \\[6pt]
    \eta_f + \tfrac{1}{2}\big(\eta_{\max} - \eta_f\big)\big(1 + \cos(\pi\, s_a(t))\big),
      & T_w < t \leq T \quad \text{(cosine anneal)}.
  \end{cases}
\end{equation}
Hence $\eta$ ramps cosine-smoothly from $\eta_0$ to $\eta_{\max}$ over the
first $5\%$ of training and is then cosine-annealed back to
$\eta_f \approx 0$ over the remaining $95\%$. The cosine-shaped warmup is
particularly well suited when continuing from a pretrained checkpoint: the
gradually increasing step size lifts the parameters out of the local optimum
they occupy without disrupting the geometry of the learned representations,
enabling stable continued training before the main annealing phase takes over.

\paragraph{Model configuration.}
The backbone consists of $N = 12$ residual blocks alternating mLSTM and sLSTM, with embedding dimension $d_{\text{model}} = 512$, $h = 4$ heads per layer, a feed-forward
expansion to dimension $d_{\text{ff}} = 2{,}048$, QK-normalization, and
dropout $p = 0.1$ applied within each block. The model operates on a context of
$L_{\text{ctx}} = 2{,}048$ time steps and produces forecasts over a horizon of
$L_{\text{pred}} = 320$ steps, with both input and output patch sizes set to
$P_{\text{in}} = P_{\text{out}} = 32$. This yields $L_{\text{ctx}}/P_{\text{in}} = 64$ input tokens and
$L_{\text{pred}}/P_{\text{out}} = 10$ output tokens per sample.

\paragraph{Training objective.}
We minimize the quantile loss over $K = 99$ equidistant quantile levels
$\{\tau_k = k/100\}_{k=1}^{99}$, augmented with soft sample-impact capping to
limit the influence of individual high-loss samples on the gradient. This
prevents pathological tail samples (e.g.\ rare regime shifts or outliers in the
synthetic mixtures) from dominating the update direction without discarding
their information content entirely.

\subsection{Long-context posttraining}
\label{app:posttraining-setup}

To adapt \tirex{-2} to longer sequences, we follow pretraining with a short
posttraining phase that extends the context length to
$L_{\text{ctx}}^{\text{pt}} = 8{,}192$ and the prediction horizon to
$L_{\text{pred}}^{\text{pt}} = 512$, while keeping the input/output patch size
fixed at $P_{\text{in}} = P_{\text{out}} = 32$.  
All architectural hyperparameters
($N$, $d_{\text{model}}$, $h$, $d_{\text{ff}}$, dropout, normalization) are
inherited unchanged from pretraining.

Posttraining is initialized from the pretraining checkpoint and runs for
$T^{\text{pt}} = 100{,}000$ optimizer steps, retaining the optimizer
configuration of the pretraining phase (AdamW with weight decay $0.01$,
gradient value-clipping at $1.0$, bf16-mixed precision) and the same training
objective (quantile loss over $K = 99$ levels with soft sample-impact capping).
The learning-rate schedule is again OneCycle but with a peak value reduced by
an order of magnitude to
$\eta_{\max}^{\text{pt}} = 1.2 \cdot 10^{-4}$, reflecting the fine-tuning
character of this phase and preventing the model from drifting away from the
pretrained solution.

While trained with a fixed prediction horizon of
$L_{\text{pred}}^{\text{pt}} = 512$, the model is not limited to it at
deployment: on dysts we forecast horizons of up to $1{,}056$ steps
(\Cref{fig:horizon_scaling_and_shift}). Likewise, the recurrent state of the
\xlstm{} backbone is not limited to the trained context length: via streaming,
we evaluate contexts of up to $32{,}000{,}000$ steps, i.e.\ $4{,}000\times$
longer than the $8{,}192$ seen in posttraining.

\begin{table}[h]
  \caption{Training hyperparameters of \tirex{-2} for the pretraining and
  long-context posttraining phases.}
  \label{tab:training-setup}
  \centering
  \small
  \begin{tabular}{lll}
    \toprule
    & Pretraining & Posttraining \\
    \midrule
    \multicolumn{3}{l}{\textit{Optimization}} \\
    \midrule
    Optimizer                       & \multicolumn{2}{l}{AdamW \citep{loshchilov_decoupled_2019}} \\
    Weight decay                    & \multicolumn{2}{l}{$0.01$} \\
    Gradient clipping (value)       & \multicolumn{2}{l}{$1.0$} \\
    Precision                       & \multicolumn{2}{l}{bf16-mixed} \\
    Hardware                        & \multicolumn{2}{l}{$2 \times$ NVIDIA H100} \\
    Total steps                     & $700{,}000$ & $100{,}000$ \\
    Effective batch size            & \multicolumn{2}{l}{$64$ (per-GPU $32$, no grad.\ accumulation)} \\
    Wall-clock time                 & $\approx 50$\,h &  \\
    Initialization                  & random &  \\
    \midrule
    \multicolumn{3}{l}{\textit{Learning-rate schedule (OneCycle)}} \\
    \midrule
    Peak LR $\eta_{\max}$           & $1.2 \cdot 10^{-3}$ & $1.2 \cdot 10^{-4}$ \\
    Warmup fraction                 & \multicolumn{2}{l}{$5\%$} \\
    Initial divisor                 & \multicolumn{2}{l}{$50$} \\
    Final divisor                   & \multicolumn{2}{l}{$10^{4}$} \\
    \midrule
    \multicolumn{3}{l}{\textit{Architecture}} \\
    \midrule
    Blocks $N$                      & \multicolumn{2}{l}{$12$ (alternating mLSTM / sLSTM)} \\
    Embedding dim $d_{\text{model}}$ & \multicolumn{2}{l}{$512$} \\
    Heads $h$                       & \multicolumn{2}{l}{$4$} \\
    FFN dim $d_{\text{ff}}$         & \multicolumn{2}{l}{$2{,}048$} \\
    Normalization                   & \multicolumn{2}{l}{QK-norm} \\
    Dropout                         & \multicolumn{2}{l}{$0.1$} \\
    Output clamp                    & \multicolumn{2}{l}{Clamp of $\mathrm{sinh}$ input to $\pm 20$ during re-scaling}\\ 
    Quantile levels $K$             & \multicolumn{2}{l}{$99$} \\
    \midrule
    \multicolumn{3}{l}{\textit{Sequence layout}} \\
    \midrule
    Context length $L_{\text{ctx}}$ & $2{,}048$ & $8{,}192$ \\
    Prediction horizon $L_{\text{pred}}$ & $320$ & $512$ \\
    Input / output patch size       & \multicolumn{2}{l}{$32$ / $32$} \\
    Input / output tokens           & $64$ / $10$ & $256$ / $16$ \\
    Streamed context length (tested) & \multicolumn{2}{l}{$32$M steps} \\
    \midrule
    \multicolumn{3}{l}{\textit{Objective}} \\
    \midrule
    Loss                            & \multicolumn{2}{l}{Quantile loss with soft sample-impact capping} \\
    Quantile levels $K$             & \multicolumn{2}{l}{$99$ (equidistant in $(0,1)$)} \\
    \bottomrule
  \end{tabular}
\end{table}

\newpage
\section{Pre-training corpus}
\label{app:pretrain-corpus}

The univariate data sources of our pre-training corpora are inherited from
\tirex{} and \tirex-1.1{}~\citep{auer_tirex_2025}. We reproduce the description here for
self-containedness and to make the boundary between the inherited univariate
sources and the multivariate extension introduced in \Cref{sec:uni2multi}
explicit. In contrast to \tirex{}, we do \emph{not} apply the TSMixup
augmentation of \citet{ansari_chronos_2024}: its role as a univariate mixing
prior is subsumed by our coupling mechanism (\Cref{sec:uni2multi}), which both
generates and mixes series under a richer set of cross-variate structures.

The univariate corpus comprises three components:

\begin{enumerate}
    \item \textbf{Chronos training data ($\sim\!30$\,M series).} We use the
    public training collection assembled by \citet{ansari_chronos_2024} as
    a source of real-world univariate \timeseries{} drawn from heterogeneous
    domains. Series are $z$-score normalized per sample and used directly,
    without TSMixup-style convex combination at this stage.

    \item \textbf{Synthetic Gaussian-process series ($\sim\!15$\,M).}
    We adopt the GP-based synthetic data pipeline of \tirex{}~\citep{auer_tirex_2025}
    verbatim, which itself is an extension of
    KernelSynth~\citep{ansari_chronos_2024}: each series is drawn from a
    zero-mean Gaussian process whose kernel is randomly composed from a fixed
    bank under $\{+, \times\}$. We refer to~\citet{auer_tirex_2025} for the
    full specification.

    \item \textbf{GIFT-Eval pre-training subset ($\sim\!2.5$\,M).}
    A subset of the pre-training corpus released alongside
    GIFT-Eval~\citep{aksu_gift-eval_2024}; concrete dataset filtering is
    handled per evaluation benchmark as described below. 
\end{enumerate}

\paragraph{Sampling protocol.}
At each training step a sample is drawn from this pool such that the Chronos component and the synthetic GP component are sampled with equal per-series probability, following the original \tirex{} pipeline.  
We additionally mix in, at a fixed rate of $1\%$, synthetic trajectories generated from
\texttt{dysts}~\citep{gilpin_chaos_2023} to
expose the model to deterministic chaotic dynamics. The corresponding
\texttt{dysts} evaluation split used in \Cref{sec:experiments} is strictly
held out and not used for training.
To form a multivariate training instance,
we draw $V \sim \cU\{1, \dots, 12\}$ univariate series from this pool
and pass them through the coupling mechanism of \Cref{sec:uni2multi}, which
imposes cross-variate dependencies on the otherwise independently sampled
series; \texttt{dysts} trajectories are exempt from this stage and enter the
batch as standalone univariate samples. The coupling mechanism thereby
both lifts the univariate marginal to a multivariate distribution and
replaces the role of TSMixup as a univariate mixing prior.

\paragraph{Per-benchmark zero-leakage corpora.}
To ensure a strictly zero-shot evaluation, we pre-train two separate model checkpoints, one for each benchmark used in \Cref{sec:experiments}: for the GIFT-Eval~\citep{aksu_gift-eval_2024} and \fevbench{}~\citep{shchur_fev-bench_2025} evaluations, we remove from the training corpus any dataset overlapping with the corresponding evaluation benchmark, following each benchmark's own leakage rules. Two datasets requiring particular attention are \texttt{chronos\_datasets/solar} and \texttt{chronos\_datasets/solar\_1h}. Following the approach of TiRex 1.0, these solar datasets do not constitute leakage for \fevbench{}, as we train at 5-minute and 1-hour resolutions while the benchmark evaluates at weekly and daily frequencies. For the GIFT-Eval checkpoint, we remove the leaking Alabama subsets of the solar datasets from the training data. The two training corpora therefore differ in their concrete dataset composition, while the sampling protocol described above is applied identically in both cases. Reported scores for each benchmark are produced by the checkpoint trained on the corpus from which that benchmark's data has been excluded. For the matched-corpus comparison in \Cref{app:add:gift-matched}, we additionally train a third checkpoint whose corpus extends the GIFT-Eval zero-shot corpus with the GIFT-Eval train splits while the test splits remain excluded.
\newpage

{\footnotesize
\begin{longtable}{@{}p{0.60\linewidth}p{0.18\linewidth}p{0.18\linewidth}@{}}
\caption{Training datasets --- \texttt{Salesforce/lotsa\_data}.}\label{tab:datasets_lotsa}\\
\toprule
\textbf{Dataset} & \textbf{TiRex-2-fev} & \textbf{TiRex-2-GIFT-Eval} \\
\midrule
\endfirsthead
\multicolumn{3}{c}{\tablename\ \thetable{} -- continued} \\
\toprule
\textbf{Dataset} & \textbf{TiRex-2-fev} & \textbf{TiRex-2-GIFT-Eval} \\
\midrule
\endhead
\midrule\multicolumn{3}{r}{\textit{Continued on next page}} \\
\endfoot
\bottomrule
\endlastfoot
\texttt{lotsa\_data/BEIJING\_SUBWAY\_30MIN} & \checkmark & \checkmark \\
\texttt{lotsa\_data/HZMETRO} & \checkmark & \checkmark \\
\texttt{lotsa\_data/LOS\_LOOP} & \checkmark & \checkmark \\
\texttt{lotsa\_data/PEMS03} & \checkmark & \checkmark \\
\texttt{lotsa\_data/PEMS04} & \checkmark & \checkmark \\
\texttt{lotsa\_data/PEMS07} & \checkmark & \checkmark \\
\texttt{lotsa\_data/PEMS08} & \checkmark & \checkmark \\
\texttt{lotsa\_data/PEMS\_BAY} & \checkmark & \checkmark \\
\texttt{lotsa\_data/Q-TRAFFIC} & \checkmark & \checkmark \\
\texttt{lotsa\_data/SHMETRO} & \checkmark & \checkmark \\
\texttt{lotsa\_data/azure\_vm\_traces\_2017} & \checkmark & --- \\
\texttt{lotsa\_data/bdg-2\_bear} & \checkmark & --- \\
\texttt{lotsa\_data/bdg-2\_fox} & \checkmark & --- \\
\texttt{lotsa\_data/bdg-2\_panther} & \checkmark & --- \\
\texttt{lotsa\_data/bdg-2\_rat} & \checkmark & --- \\
\texttt{lotsa\_data/beijing\_air\_quality} & \checkmark & \checkmark \\
\texttt{lotsa\_data/borealis} & \checkmark & \checkmark \\
\texttt{lotsa\_data/borg\_cluster\_data\_2011} & \checkmark & --- \\
\texttt{lotsa\_data/buildings\_900k} & \checkmark & \checkmark \\
\texttt{lotsa\_data/bull} & --- & \checkmark \\
\texttt{lotsa\_data/cdc\_fluview\_ilinet} & \checkmark & \checkmark \\
\texttt{lotsa\_data/cdc\_fluview\_who\_nrevss} & \checkmark & \checkmark \\
\texttt{lotsa\_data/china\_air\_quality} & \checkmark & \checkmark \\
\texttt{lotsa\_data/cockatoo} & --- & \checkmark \\
\texttt{lotsa\_data/covid19\_energy} & --- & \checkmark \\
\texttt{lotsa\_data/covid\_mobility} & --- & \checkmark \\
\texttt{lotsa\_data/elf} & --- & \checkmark \\
\texttt{lotsa\_data/extended\_web\_traffic\_with\_missing} & --- & \checkmark \\
\texttt{lotsa\_data/favorita\_sales} & \checkmark & \checkmark \\
\texttt{lotsa\_data/favorita\_transactions} & --- & \checkmark \\
\texttt{lotsa\_data/gfc12\_load} & --- & \checkmark \\
\texttt{lotsa\_data/gfc14\_load} & --- & \checkmark \\
\texttt{lotsa\_data/gfc17\_load} & --- & \checkmark \\
\texttt{lotsa\_data/godaddy} & \checkmark & \checkmark \\
\texttt{lotsa\_data/hog} & --- & \checkmark \\
\texttt{lotsa\_data/ideal} & \checkmark & \checkmark \\
\texttt{lotsa\_data/kaggle\_web\_traffic\_weekly} & \checkmark & \checkmark \\
\texttt{lotsa\_data/kdd2022} & --- & \checkmark \\
\texttt{lotsa\_data/largest\_2017} & \checkmark & \checkmark \\
\texttt{lotsa\_data/largest\_2018} & \checkmark & \checkmark \\
\texttt{lotsa\_data/largest\_2019} & \checkmark & \checkmark \\
\texttt{lotsa\_data/largest\_2020} & \checkmark & \checkmark \\
\texttt{lotsa\_data/largest\_2021} & \checkmark & \checkmark \\
\texttt{lotsa\_data/lcl} & \checkmark & --- \\
\texttt{lotsa\_data/london\_smart\_meters\_with\_missing} & --- & \checkmark \\
\texttt{lotsa\_data/monash\_m3\_other} & --- & \checkmark \\
\texttt{lotsa\_data/oikolab\_weather} & --- & \checkmark \\
\texttt{lotsa\_data/pdb} & --- & \checkmark \\
\texttt{lotsa\_data/pedestrian\_counts} & --- & \checkmark \\
\texttt{lotsa\_data/project\_tycho} & --- & \checkmark \\
\texttt{lotsa\_data/residential\_load\_power} & \checkmark & \checkmark \\
\texttt{lotsa\_data/residential\_pv\_power} & \checkmark & \checkmark \\
\texttt{lotsa\_data/sceaux} & \checkmark & \checkmark \\
\texttt{lotsa\_data/smart} & \checkmark & \checkmark \\
\texttt{lotsa\_data/solar\_power} & \checkmark & \checkmark \\
\texttt{lotsa\_data/spain} & --- & \checkmark \\
\texttt{lotsa\_data/subseasonal} & \checkmark & \checkmark \\
\texttt{lotsa\_data/subseasonal\_precip} & \checkmark & \checkmark \\
\texttt{lotsa\_data/sunspot\_with\_missing} & --- & \checkmark \\
\texttt{lotsa\_data/tourism\_monthly} & --- & \checkmark \\
\texttt{lotsa\_data/vehicle\_trips\_with\_missing} & --- & \checkmark \\
\texttt{lotsa\_data/wiki-rolling\_nips} & --- & \checkmark \\
\texttt{lotsa\_data/wind\_power} & \checkmark & \checkmark \\
\end{longtable}
}

{\footnotesize
\begin{longtable}{@{}p{0.60\linewidth}p{0.18\linewidth}p{0.18\linewidth}@{}}
\caption{Training datasets --- \texttt{autogluon/chronos\_datasets}.}\label{tab:datasets_chronos}\\
\toprule
\textbf{Dataset} & \textbf{TiRex-2-fev} & \textbf{TiRex-2-GIFT-Eval} \\
\midrule
\endfirsthead
\multicolumn{3}{c}{\tablename\ \thetable{} -- continued} \\
\toprule
\textbf{Dataset} & \textbf{TiRex-2-fev} & \textbf{TiRex-2-GIFT-Eval} \\
\midrule
\endhead
\midrule\multicolumn{3}{r}{\textit{Continued on next page}} \\
\endfoot
\bottomrule
\endlastfoot
\texttt{chronos\_datasets\_extra/brazilian\_cities\_temperature} & \checkmark & \checkmark \\
\texttt{chronos\_datasets/dominick} & --- & \checkmark \\
\texttt{chronos\_datasets/electricity\_15min} & \checkmark & --- \\
\texttt{chronos\_datasets/exchange\_rate} & --- & \checkmark \\
\texttt{chronos\_datasets/m4\_daily} & \checkmark & --- \\
\texttt{chronos\_datasets/m4\_hourly} & \checkmark & --- \\
\texttt{chronos\_datasets/m4\_monthly} & \checkmark & --- \\
\texttt{chronos\_datasets/m4\_weekly} & \checkmark & --- \\
\texttt{chronos\_datasets/mexico\_city\_bikes} & \checkmark & \checkmark \\
\texttt{chronos\_datasets/monash\_australian\_electricity} & --- & \checkmark \\
\texttt{chronos\_datasets/monash\_cif\_2016} & --- & \checkmark \\
\texttt{chronos\_datasets/monash\_electricity\_hourly} & \checkmark & --- \\
\texttt{chronos\_datasets/monash\_electricity\_weekly} & \checkmark & --- \\
\texttt{chronos\_datasets/monash\_fred\_md} & --- & \checkmark \\
\texttt{chronos\_datasets/monash\_kdd\_cup\_2018} & \checkmark & --- \\
\texttt{chronos\_datasets/monash\_london\_smart\_meters} & \checkmark & \checkmark \\
\texttt{chronos\_datasets/monash\_m1\_monthly} & --- & \checkmark \\
\texttt{chronos\_datasets/monash\_m1\_quarterly} & --- & \checkmark \\
\texttt{chronos\_datasets/monash\_m1\_yearly} & --- & \checkmark \\
\texttt{chronos\_datasets/monash\_m3\_monthly} & --- & \checkmark \\
\texttt{chronos\_datasets/monash\_m3\_quarterly} & --- & \checkmark \\
\texttt{chronos\_datasets/monash\_m3\_yearly} & --- & \checkmark \\
\texttt{chronos\_datasets/monash\_nn5\_weekly} & --- & \checkmark \\
\texttt{chronos\_datasets/monash\_pedestrian\_counts} & \checkmark & \checkmark \\
\texttt{chronos\_datasets/monash\_rideshare} & \checkmark & \checkmark \\
\texttt{chronos\_datasets/monash\_temperature\_rain} & \checkmark & --- \\
\texttt{chronos\_datasets/monash\_tourism\_quarterly} & --- & \checkmark \\
\texttt{chronos\_datasets/monash\_tourism\_yearly} & --- & \checkmark \\
\texttt{chronos\_datasets/monash\_traffic} & --- & \checkmark \\
\texttt{chronos\_datasets/monash\_weather} & --- & \checkmark \\
\texttt{chronos\_datasets/nn5} & --- & \checkmark \\
\texttt{chronos\_datasets/solar} & \checkmark & \checkmark \\
\texttt{chronos\_datasets/solar\_1h} & \checkmark & \checkmark \\
\texttt{chronos\_datasets\_extra/spanish\_energy\_and\_weather} & \checkmark & \checkmark \\
\texttt{chronos\_datasets/taxi\_1h} & \checkmark & \checkmark \\
\texttt{chronos\_datasets/taxi\_30min} & \checkmark & \checkmark \\
\texttt{chronos\_datasets/uber\_tlc\_daily} & \checkmark & \checkmark \\
\texttt{chronos\_datasets/uber\_tlc\_hourly} & \checkmark & \checkmark \\
\texttt{chronos\_datasets/ushcn\_daily} & \checkmark & \checkmark \\
\texttt{chronos\_datasets/weatherbench\_daily} & \checkmark & \checkmark \\
\texttt{chronos\_datasets/weatherbench\_hourly\_*} (11 vars) & \checkmark & \checkmark \\
\texttt{chronos\_datasets/weatherbench\_weekly} & \checkmark & \checkmark \\
\texttt{chronos\_datasets/wiki\_daily\_100k} & \checkmark & \checkmark \\
\texttt{chronos\_datasets/wind\_farms\_daily} & \checkmark & \checkmark \\
\texttt{chronos\_datasets/wind\_farms\_hourly} & \checkmark & \checkmark \\
\end{longtable}
}

\newpage
\section{Synthetic multivariate coupling: background and design}
\label{app:coupler}

This appendix provides the conceptual background for the synthetic
coupling pipeline introduced in \Cref{sec:uni2multi}. We specify the
phenomena the pipeline is designed to cover during training of
\tirex{-2}, the rationale for their inclusion, and the design
principles underlying the construction.

\paragraph{Motivation.}
Real multivariate \timeseries{} rarely arise from a single generative
process. They typically combine several qualitatively distinct sources
of cross-variate structure: shared unobserved drivers, lagged
dependencies, deterministic functional relationships between variates,
and joint stochastic trends. Observational pipelines superimpose
further structure (irregular sampling, partial observability, sensor
dropouts, quantisation, and asynchronous updates) that is generally
inseparable from the underlying signal. A foundation model intended
for zero-shot generalisation across domains must therefore handle all
of these regimes simultaneously, since the dominant regime is
typically unknown a priori and may vary within a single dataset.

Curated multivariate corpora are insufficient to enforce this breadth,
whereas large univariate corpora are abundant. We therefore construct
multivariate training examples on the fly from the univariate pool by
sampling from a broad menu of cross-variate dependency types, combined
with a rich set of observational artefacts. The pipeline is not
intended to replicate any specific real-world dataset; rather, it
ensures that no single inductive bias dominates, so that shared latent
factors, lagged causal influence, deterministic covariate
relationships, and common stochastic trends are each represented with
non-negligible probability in the training distribution.

\paragraph{Design principles.}
The pipeline is governed by three principles. \emph{Coverage}: the
training distribution spans qualitatively distinct dependency types,
including indirect (latent-factor) structure, directed lagged
causation, and direct pointwise functional relationships.
\emph{Compositionality}: each stage is independently randomised per
example, yielding a combinatorial enlargement of the effective
training distribution relative to any enumeration of fixed scenarios.
\emph{Observational realism}: the data presented to \tirex{-2}
reflects the structural artefacts encountered in applied forecasting,
which frequently account for the gap between benchmark and deployment
performance.

\paragraph{Pipeline overview.}
The pipeline comprises three stages, each randomised per example. The
first stage diversifies the marginal behaviour of individual
univariate series (amplitude profile, dynamic range, and the
presence and shape of localised events) so that the joint structure
imposed in subsequent stages is not confounded with marginal
variability. The second stage introduces cross-variate dependencies
by sampling from a collection of coupling mechanisms, each targeting
a distinct region of dependency space. The third stage applies a
sequence of randomised observational transforms. The mechanism within
each stage is itself sampled, so that the joint distribution over
training examples is induced by a procedural prior rather than by a
single parametric model.

\paragraph{Coverage of dependency phenomena.}
The coupling mechanisms in the second stage are selected to span
complementary forms of multivariate structure. Each constitutes a
stylised representative of a class of phenomena; the equations from
\Cref{sec:uni2multi} are restated below to fix notation.

\emph{Indirect coupling through linear mixing,}
\begin{equation*}
  \Bx_t \;=\; \BA\,\Bu_t,
\end{equation*}
captures the case in which observed series arise as combinations of
a smaller number of underlying drivers. The induced correlation
structure varies continuously with the spectrum of $\BA$,
ranging from near-independence to near-collinearity. We sample from
several qualitatively distinct spectral regimes to avoid implicit
specialisation to any particular effective rank or condition number.

\emph{Directed, lagged influence} is represented by structural causal
models over a randomly sampled directed acyclic graph. In the linear
case,
\begin{equation*}
  x_{f,t} \;=\; \sum_{i\in\mathrm{pa}(f)} \alpha_{if}\,u_{i,\,t-\tau_{if}}
  \;+\; \varepsilon_{f,t},
\end{equation*}
this class introduces dependencies that instantaneous mixing cannot
produce, in particular a temporal asymmetry between cause and effect.
The nonlinear extension,
\begin{equation*}
  x_{f,t} \;=\; h\!\bigl(u_{k,\,t-\tau_k}\bigr)
  \,\sum_{i\in\mathrm{pa}(f)} g_{if}\!\bigl(u_{i,\,t-\tau_{if}}\bigr),
\end{equation*}
adds a multiplicative modulation component, which serves as a
proxy for threshold-driven and regime-switching dynamics. Such
dynamics are characteristic of gated systems and of many physical and
economic processes in which one variable controls the activity of
another. The pipeline does not commit to a parametric form for the
modulating function $h$ or the edge-level nonlinearities $g_{if}$.

\emph{Common stochastic trends} are represented by a cointegration
mechanism,
\begin{equation*}
  \Bx_t \;=\; \BLa\,\Bs_t
  \;+\; \Bxi_t,
\end{equation*}
with non-stationary shared drivers $\Bs$ and stationary
deviations $\Bxi$. The relevant phenomenon is that
individual series may drift without bound while specific linear
combinations remain stationary, a regime that is poorly approximated
by either linear mixing or SCM-style coupling.

\emph{Direct functional coupling,}
\begin{equation*}
  x_{f,t} \;=\; \phi_f\!\bigl(u_{0,t}\bigr) + \varepsilon_{f,t},
\end{equation*}
represents the opposite extreme: a covariate is a deterministic
transformation of the target up to additive noise. This regime
corresponds to calendar features, derived quantities, and sensors
that observe a common underlying signal through distinct
nonlinearities, and is the most informative for the forecasting
target. It complements the indirect dependency classes by exposing
the model to covariates carrying near-deterministic information,
which is the regime the asymmetric variate mixer is designed to
exploit.

The identity and univariate pass-through case,
\begin{equation*}
  x_{f,t} \;=\; u_{f,t},
\end{equation*}
is included explicitly to preserve univariate forecasting performance.
In its absence, exposure to coupled data biases the model toward
assuming cross-variate structure even when none is present, which is
particularly detrimental in univariate mode, where the variate mixer
is bypassed.

\paragraph{Observational layer.}
The third stage addresses the empirical observation that the gap
between a well-specified generative process and the observed data is
frequently dominated by observational artefacts. Variates may be
reordered arbitrarily across datasets, sampled asynchronously, missing
in contiguous blocks (due to joint blackouts or independent sensor
faults), observed only up to the forecast origin rather than over the
full horizon $O$, or discretised in value or time. Each of these
artefacts must be handled by a deployed forecasting model and is
absent from clean synthetic data. The observational layer is therefore
treated as a first-class component of the pipeline. The artefact
families it covers (variate permutation, smooth time warping, patch
masking generalising the contiguous-patch scheme of
\tirex{}~\citep{auer_tirex_2025}, partial future observability for a
random subset of future covariates, and value- and
time-discretisation) were selected on the basis of artefact patterns
prevalent in applied forecasting. Partial future observability
prevents \tirex{-2} from becoming dependent on the future-covariate
channel being fully populated, training it to exploit such
information when available without conditioning on its presence.

\newpage
\section{Evaluation Metrics}
\label{app:eval-metrics}

We assess both point and probabilistic forecast accuracy. Point accuracy is
measured with the Mean Absolute Scaled Error (MASE) on both \fevbench{} and
GIFT-Eval. Probabilistic accuracy is measured with the Continuous Ranked
Probability Score (CRPS) on GIFT-Eval and with the Scaled Quantile Loss (SQL)
on \fevbench{}, following the protocol of each
benchmark~\citep{aksu_gift-eval_2024,shchur_fev-bench_2025}. All metrics are
computed per target series over the forecast horizon and then aggregated
according to the respective benchmark protocol.

Throughout, let $y_t$ denote the observed value and $\hat{y}_t$ the point
forecast at \timestep{} $t$, and let $\hat{y}^{(q)}_t$ denote the predicted
$q$-quantile for $q\in\mathcal{Q}=\{0.1,0.2,\dots,0.9\}$. Forecasts span the
horizon $t=T+1,\dots,T+O$, and $s$ is the seasonal period implied by the data
frequency. We denote the historical seasonal-naive error by
\begin{equation*}
  a \;=\; \frac{1}{T-s}\sum_{t=s+1}^{T}\bigl|y_t - y_{t-s}\bigr|
\end{equation*}

and the quantile (pinball) loss at level $q$ by
\begin{equation*}
  \rho_q\!\left(y_t,\hat{y}^{(q)}_t\right) \;=\;
  q\,\bigl(y_t-\hat{y}^{(q)}_t\bigr)_+ + (1-q)\,\bigl(\hat{y}^{(q)}_t-y_t\bigr)_+,
  \qquad (z)_+ = \max(z,0).
\end{equation*}

\paragraph{Mean Absolute Scaled Error (MASE).}
MASE scales the mean absolute forecast error by the in-sample seasonal-naive
error $a$, yielding a scale-free point metric:
\begin{equation*}
  \mathrm{MASE} \;=\; \frac{1}{O\,a}\sum_{t=T+1}^{T+O}\bigl|y_t-\hat{y}_t\bigr|.
\end{equation*}

\paragraph{Continuous Ranked Probability Score (CRPS).}
CRPS measures the squared distance between the predictive CDF $F_t$ and the
observation,
\begin{equation*}
  \mathrm{CRPS} \;=\; \frac{1}{O}\sum_{t=T+1}^{T+O}
  \int_{-\infty}^{\infty}\bigl(F_t(u)-\mathbf{1}\{y_t\le u\}\bigr)^2\,\mathrm{d}u,
\end{equation*}
which we approximate by the mean weighted quantile loss over
$\mathcal{Q}$~\citep{aksu_gift-eval_2024}:
\begin{equation*}
  \mathrm{CRPS} \;\approx\; \frac{1}{|\mathcal{Q}|}\sum_{q\in\mathcal{Q}}
  \frac{2\sum_{t=T+1}^{T+O}\rho_q\!\left(y_t,\hat{y}^{(q)}_t\right)}
       {\sum_{t=T+1}^{T+O}\bigl|y_t\bigr|}.
\end{equation*}

\paragraph{Scaled Quantile Loss (SQL).}
SQL is the probabilistic analogue of MASE: it aggregates the quantile loss over
$\mathcal{Q}$ and normalizes by the seasonal-naive error $a$ rather than by
$\sum_t|y_t|$, keeping the metric scale-free~\citep{shchur_fev-bench_2025}:
\begin{equation*}
  \mathrm{SQL} \;=\; \frac{2}{O\,a}\sum_{t=T+1}^{T+O}\sum_{q\in\mathcal{Q}}
  \rho_q\!\left(y_t,\hat{y}^{(q)}_t\right).
\end{equation*}

\paragraph{Aggregation across tasks.}
On \fevbench{} we aggregate per-task performance into two complementary
marginal statistics, following~\citet{shchur_fev-bench_2025}. Let $E_{rj}$
denote the error (SQL) of model $j$ on task $r$, over $R$ tasks and $M$ models.

The \emph{average win rate} $W_j$ is the probability that model $j$ achieves a
lower error than another randomly chosen model $k\neq j$ on a randomly chosen
task, with ties counted as half a win:
\begin{equation*}
  W_j \;=\; \frac{1}{R\,(M-1)} \sum_{r=1}^{R} \sum_{\substack{k=1\\k\neq j}}^{M}
  \Bigl[\mathbf{1}(E_{rj}<E_{rk}) + \tfrac{1}{2}\,\mathbf{1}(E_{rj}=E_{rk})\Bigr],
\end{equation*}
ranging from $0$ (worst) to $1$ (best).

The \emph{skill score} $S_j$ quantifies the average relative error reduction of
model $j$ against a fixed baseline $\beta$ (Seasonal Naive), aggregated as a
geometric mean of per-task error ratios:
\begin{equation*}
  S_j \;=\; 1 - \sqrt[\leftroot{-1}\uproot{2}R]{\prod_{r=1}^{R}
  \operatorname{clip}\!\left(\frac{E_{rj}}{E_{r\beta}};\,\ell,\,u\right)},
  \qquad
  \operatorname{clip}(x;\ell,u)=\max(\ell,\min(x,u)),
\end{equation*}
with clipping bounds $\ell=10^{-2}$, $u=10^{2}$ to bound the influence of
extreme ratios. Positive values indicate better-than-baseline performance.

\section{Code and model checkpoints}
\label{app:code}

The code repository for the model is hosted on GitHub:
\url{https://github.com/NX-AI/tirex-2}.
The pretrained checkpoints are available on Hugging Face
(\url{https://huggingface.co/NX-AI/TiRex-2}), one per training corpus of
\Cref{app:pretrain-corpus}:

\begin{itemize}
  \item \textbf{GIFT-Eval, zero-shot}
    (\href{https://huggingface.co/NX-AI/TiRex-2-gifteval-zs}{\texttt{NX-AI/TiRex-2-gifteval-zs}}):
    excludes any overlap with the GIFT-Eval pretraining and evaluation
    datasets from pretraining. Used for the GIFT-Eval results of
    \Cref{sec:experiments}.
  \item \textbf{GIFT-Eval, matched corpus}
    (\href{https://huggingface.co/NX-AI/TiRex-2-gifteval-pretrain}{\texttt{NX-AI/TiRex-2-gifteval-pretrain}}):
    adds the GIFT-Eval pretraining collection to the zero-shot corpus. Used
    for the comparison with Chronos-2 in \Cref{app:add:gift-matched}.
  \item \textbf{\fevbench{}, zero-shot}
    (\href{https://huggingface.co/NX-AI/TiRex-2-fevbench}{\texttt{NX-AI/TiRex-2-fevbench}}):
    excludes all \fevbench{} evaluation datasets from pretraining, following
    the same approach as for GIFT-Eval. Used for all \fevbench{} results.
\end{itemize}

\end{document}